\documentclass[graybox]{svmult}
\pdfoutput=1  %
\let\svthefootnote\thefootnote

\usepackage{helvet}         %

\usepackage{makeidx}         %
\usepackage{graphicx}        %
\usepackage{multicol}        %
\usepackage[bottom]{footmisc}%

\usepackage{amsmath,amssymb} %
\usepackage{subfig}
\usepackage{algorithm}
\usepackage{algpseudocode}
\usepackage{epstopdf}
\usepackage[numbers]{natbib}
\newcommand{\mb}[1]{\mathbf{#1}}
\newcommand{\bs}[1]{\boldsymbol{#1}}
\newcommand{\N}[0]{\mathcal{N}}
\newcommand{\etal}{\textit{et al}.}

\begin{document}
\title*{Bayesian Neural Networks: An Introduction and Survey}
\author{Ethan Goan, Clinton Fookes}
\institute{Ethan Goan \at Queensland University of Technology \email{ej.goan@qut.edu.au}}
\maketitle
\abstract*{Neural Networks (NNs) have provided state-of-the-art results for many challenging machine learning tasks such as detection, regression and classification across the domains of computer vision, speech recognition and natural language processing. Despite their success, they are often implemented in a frequentist scheme, meaning they are unable to reason about uncertainty in their predictions. This article introduces Bayesian Neural Networks (BNNs) and the seminal research regarding their implementation. Different approximate inference methods are compared, and used to highlight where future research can improve on current methods.}

\abstract{Neural Networks (NNs) have provided state-of-the-art results for many challenging machine learning tasks such as detection, regression and classification across the domains of computer vision, speech recognition and natural language processing. Despite their success, they are often implemented in a frequentist scheme, meaning they are unable to reason about uncertainty in their predictions. This article introduces Bayesian Neural Networks (BNNs) and the seminal research regarding their implementation. Different approximate inference methods are compared, and used to highlight where future research can improve on current methods.}

\let\thefootnote\relax\footnote{\textit{Case Studies in Applied Data Science}, (Eds K Mengersen, P Pudlo and CP Robert), Lecture Notes in Mathematics, Springer 2020 \\}
\addtocounter{footnote}{-1}\let\thefootnote\svthefootnote
\section{Introduction}
\label{sec:intro}
Biomimicry has long served as a basis for technological
developments. Scientists and engineers have repeatedly used
knowledge of the physical world to emulate nature's elegant solutions
to complex problems which have evolved over billions of years. An
important example of biomimicry in statistics and machine learning has
been the development of the perceptron \cite{rosenblatt1958}, which
proposes a mathematical model based on the physiology of a neuron.
The machine learning community has used this
concept\footnote{While also relaxing many of the constraints imposed
  by a physical model of a natural neuron \cite{bishop2006}. It should be emphasised that there is very little evidence to suggest that that arrangement of neurons seen in NNs are an accurate model of any physiological brain.} to develop
statistical models of highly interconnected arrays of neurons to
create Neural Networks (NNs).
\par
Though the concept of NNs has been known for many decades, it is only
recently that applications of these network have seen such
prominence. The lull in research and development for NNs was largely
due to three key factors: lack of sufficient algorithms to train these
networks, the large amount of data required to train complex networks
and the large amount of computing resources required during the training
process. In 1986, \cite{rumelhart1986learning} introduced the backpropagation algorithm
to address the problem of
efficient training for these networks. Though an efficient means of
training was available, considerable compute resources was still
required for the ever increasing size of new networks. This problem
was addressed in \cite{oh2004gpu, claudiu2010, krizhevsky2012imagenet}
where it was shown that general purpose GPUs could be used to
efficiently perform many of the operations required for training. As
digital hardware continued to advance, the number of sensors able to
capture and store real world data increased. With
efficient training methods, improved computational resources and large
data sets, training of complex NNs became truly feasible.
\par
In the vast majority of cases, NNs are used within a frequentist
perspective; using available data, a user defines a network architecture and cost function, which is then optimised to
allow us to gain point estimate predictions. Problems arise from
this interpretation of NNs. Increasing the number of parameters (often
called weights in machine learning literature), or the depth of the
model increases the capacity of the network, allowing it to represent
functions with greater non-linearities. This increase in capacity
allows for more complex tasks to be addressed with NNs, though when frequentist methodologies are applied, leaves
them highly prone to overfitting to the training data. The use of
large data sets and regularisation methods such as finding a MAP
estimate can limit the complexity of functions learnt by the
networks and aid in avoiding overfitting.
\par
Neural Networks have provided state-of-the-art results for numerous machine learning and Artificial intelligence (AI) applications, such as image classification \cite{krizhevsky2012imagenet, simonyan2014very, szegedy2015going}, object detection \cite{girshick2014rich, ren2015faster, redmon2016you} and speech recognition \cite{mohamed2012acoustic, dahl2012context, hinton2012deep, pmlr-v48-amodei16}. Other networks such as the AlphaGo model developed by DeepMind \cite{silver2017mastering} have emphasised the potential of NNs for developing AI systems, garnering a wide audience interested in the development of these networks.
As the performance of NNs has continued to increase, the interest in their
development and adoption by certain industries becomes more
prominent. NNs are currently used in manufacturing
\cite{mckinsey2017}, asset management \cite{van2017} and human
interaction technologies \cite{oord2017, siri2017}.
\par
Since the deployment of NNs in industry, there have been a number of incidents where failings in these systems has led to models acting unethically and unsafely. This includes models demonstrating considerable gender and racial bias against marginalised groups\cite{wakefield2016, Guynn2015, buolamwini2018gender} or to more extreme cases resulting in loss of life\cite{tesla2016, abc2018}. NNs are a statistical black-box models, meaning that the decision process is not based on a well-defined and intuitive protocol. Instead decisions are made in an uninterpretable manner, with hopes that the reasonable decisions will be made based on previous evidence provided in training data\footnote{Due to this black-box nature, the performance of these models is justified entirely through empirical means.}. As such, the
implementation of these systems in social and safety critical
environments raises considerable ethical concerns. The European Union
released a new regulation\footnote{This regulation came into effect
on the 25th of May, 2018 across the EU \cite{eu2016}.} which
effectively states that users have a ``right to an explanation''
regarding decisions made by AI systems \cite{eu2016,
goodman2016european}. Without clear understanding of their operation
or principled methods for their design, experts from other domains
remain apprehensive about the adoption of current technology
\cite{vu2018, holzinger2017we, caruana2015intelligible}. These limitations have motivated research efforts into the field of Explainable AI \cite{gunning2017explainable}.
\par
Adequate
engineering of NNs requires a sound understanding of their capabilities and limitations; to identify their shortcomings prior to deployment as apposed
to the current practice of investigating these limitations in the wake
of these tragedies. With NNs being a statistical black-box, interpretation and explanation of the decision making process eludes current
theory. This lack of interpretation and over-confident estimates
provided by the frequentist perspective of common NNs makes them
unsuitable for high risk domains such as medical diagnostics and autonomous vehicles. Bayesian statistics offers natural way to reason about uncertainty in predictions, and can provide insight into how these decisions are made.
\par
\begin{figure}[!h]
  \centering
  \subfloat[][]{\includegraphics[width=0.5\textwidth]{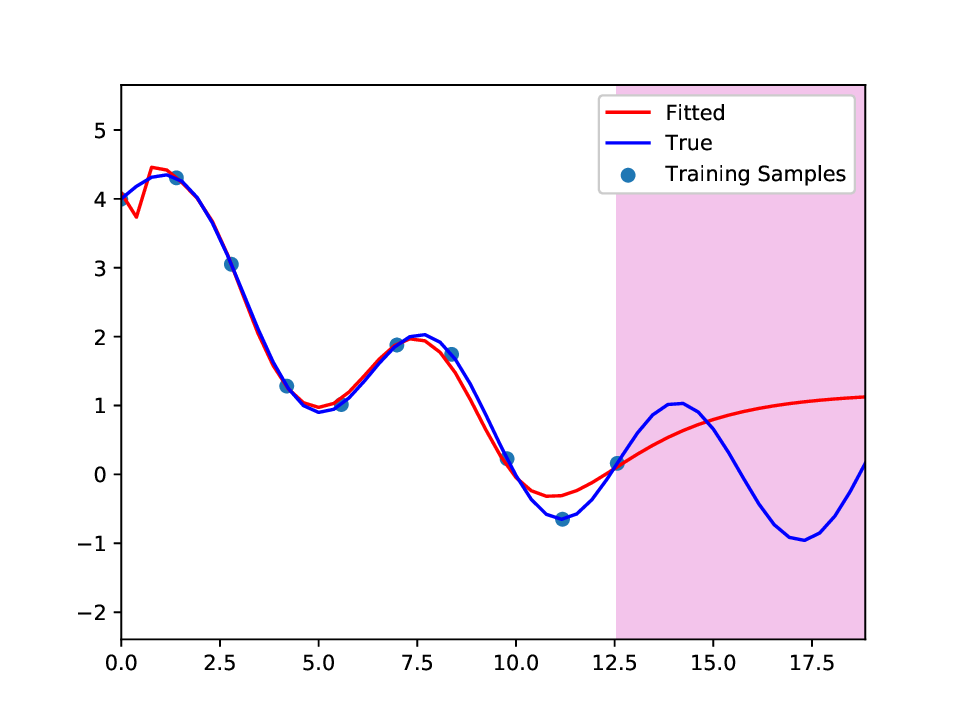}}
  \subfloat[][]{\includegraphics[width=0.5\textwidth]{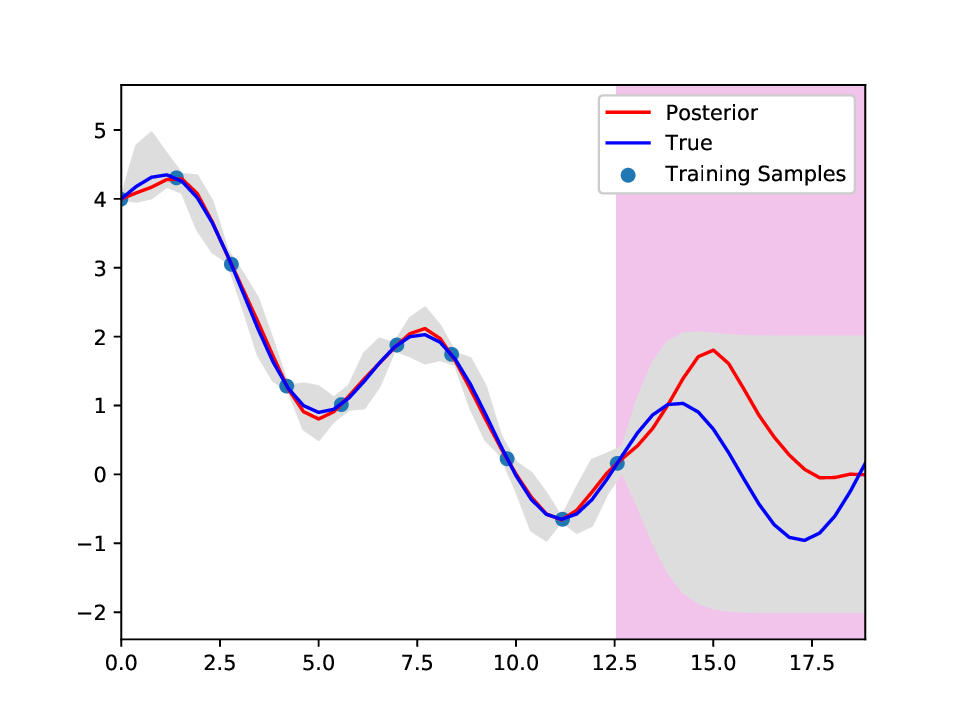}}
  \caption{Comparison of neural network to traditional probabilistic
    methods for a regression task, with no training data in the purple
    region. (a) Regression output using a neural network with 2 hidden layers; (b) Regression using a Gaussian Process framework, with
    grey bar representing $\pm 2$ std. from expected value.}
  \label{fig:regression}
\end{figure}

Figure \ref{fig:regression} compares Bayesian methods for performing
regression with that of a simple neural network, and illustrates the
importance of measuring uncertainty. While both methods perform well
within the bounds of the training data, where extrapolation is
required, the probabilistic method provides a full distribution of the
function output as opposed to the point estimates provided by the
NN. The distribution over outputs provided by
probabilistic methods allows for the development of trustworthy models, in
that they can identify uncertainty in a prediction. Given that NNs are
the most promising model for generating AI systems, it is important
that we can similarly trust their predictions.
\par
A Bayesian perspective allows us to address many of the challenges
currently faced within NNs. To do this, a distribution is placed over
the network parameters, and the resulting network is then termed a
Bayesian Neural Network (BNN). The goal of a BNN is to have a model of
high capacity that exhibits the important theoretical benefits of
Bayesian analysis. Recent research has investigated how Bayesian
approximations can be applied to NNs in practice. The challenge with these
methods is deploying models that provide accurate predictions within reasonable computation
constraints\footnote{The term ``reasonable'' largely depends on the
  context. Many neural networks are currently trained using some of the
  largest computing facilities available, containing thousands of GPU
  devices.}.
\par
This document aims to provide an accessible introduction to BNNs, accompanied by a survey of seminal works in the field and experiments to motivate discussion into the capabilities and limits of current methods. A survey of all research items across the Bayesian and machine learning literature related to BNNs could fill multiple text books. As a result, items included in this survey only intend to inform the reader on the overarching narrative that has motivated their research. Similarly, derivations of many of they key results have been omitted, with the final result being listed accompanied by reference to the original source. Readers inspired by this exciting research area are encouraged to consult prior surveys: \cite{mackay1995probable} which surveys the early developments in BNNs, \cite{lampinen2001bayesian} which discusses the specifics of a full Bayesian treatment for NNs, and~\cite{wang2016towards} which surveys applications of approximate Bayesian inference to modern network architectures.
\par
This document should be suitable for all in the statistics field, though the primary audience of interest are those more familiar with machine learning concepts. Despite seminal references for new machine learning scholars almost equivalently being Bayesian texts~\cite{bishop2006, murphey2012machine}, in practice there has been a divergence between much of the modern machine learning and Bayesian statistics research. It is hoped that this survey will help highlight similarities between some modern research in BNNs and statistics, to emphasis the importance of a probabilistic perspective within machine learning and to encourage future collaboration/unison between the machine learning and statistics fields.

\section{Literature Survey}
\label{sec:lit_review}
\subsection{Neural Networks}
\begin{figure}[!h]
  \centering
  \includegraphics[width=0.6\linewidth]{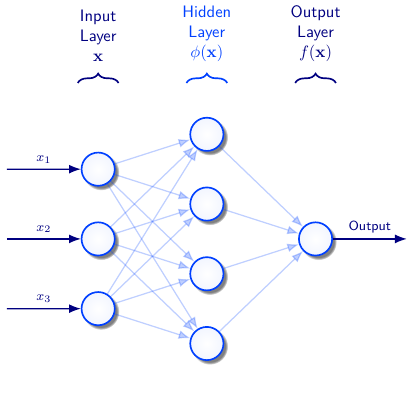}
  \caption{Example of a NN architecture with a single hidden layer for
    either binary classification or 1-D regression. Each node
    represents a neuron or a state where the summation and activation
    of input states is performed. Arrows are the parameters
    (weights) indicating the strength of connection between neurons.}
  \label{fig:mlp}
\end{figure}

Before discussing a Bayesian perspective of NNs, it is important to
briefly survey the fundamentals of neural computation and to define
the notation to be used throughout the chapter. This survey will focus on the
primary network structure of interest, the Multi-Layer Perceptron
(MLP) network. The MLP serves as the basis for NNs, with modern
architectures such as convolutional networks having an equivalent MLP
representation. Figure~\ref{fig:mlp} illustrates a simple MLP with a single hidden layer suitable for regression or
classification. For this
network with an input $\mb{x}$ of dimension $N_1$, the output of the $\mb{f}$ network can be
modelled as,
\begin{equation}
  \label{eq:basis}
  \phi_j = \sum_{i = 1}^{N_1} a(x_i w^1_{ij}),
\end{equation}
\begin{equation}
  \label{eq:mlp}
  f_k = \sum_{j = 1}^{N_2} g(\phi_j w^2_{jk}).
\end{equation}
The parameters $w$ represent the weighted connection between neurons from subsequent layers, and the superscripts denoting the layer number. Equation~\ref{eq:basis} represents the output of the hidden layer, which will be of dimension $N_2$. The $k^{th}$ output of the network is then a summation over the $N_2$ outputs from the prior hidden layer. This modelling scheme can be expanded to include many hidden layers, with the input of each layer being the output of the layer immediately prior. A bias
value is often added during each layer, though is omitted throughout this chapter in favour of simplicity.
\par
\begin{figure}[!h]
  \centering
  \subfloat[Sigmoid][]{\includegraphics[width=0.5\textwidth]{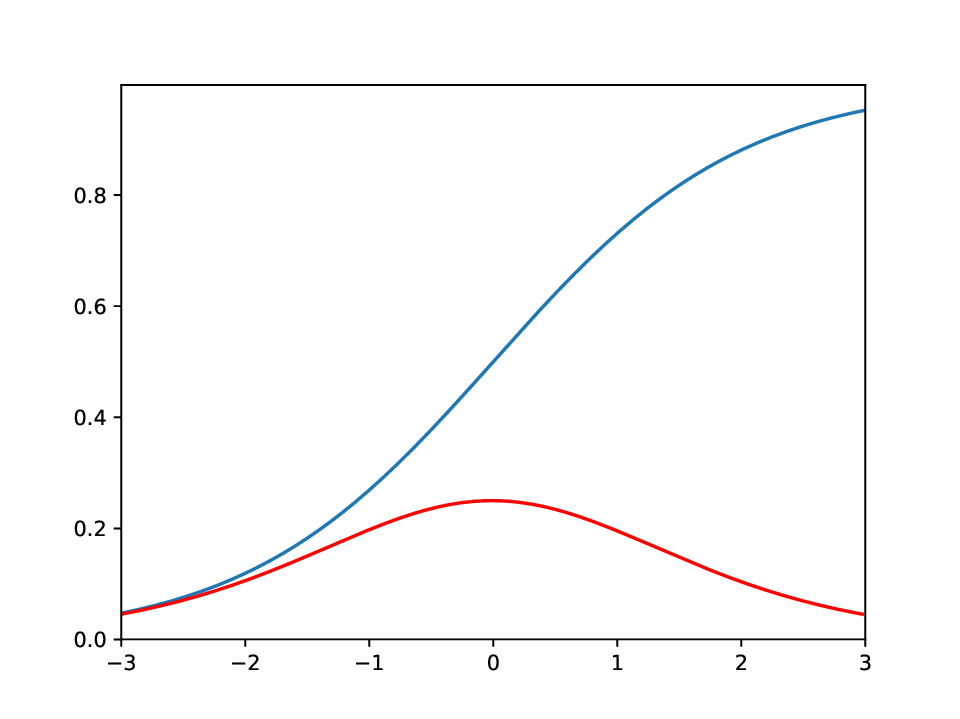}}
  \subfloat[TanH][]{\includegraphics[width=0.5\textwidth]{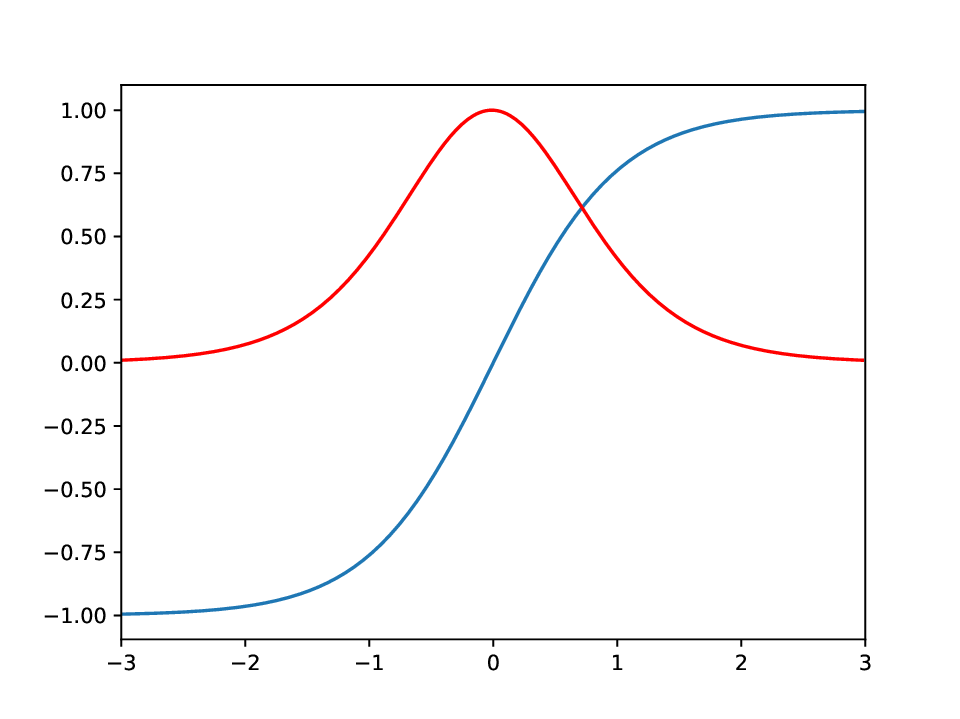}}
  \\
  \subfloat[ReLU][]{\includegraphics[width=0.5\textwidth]{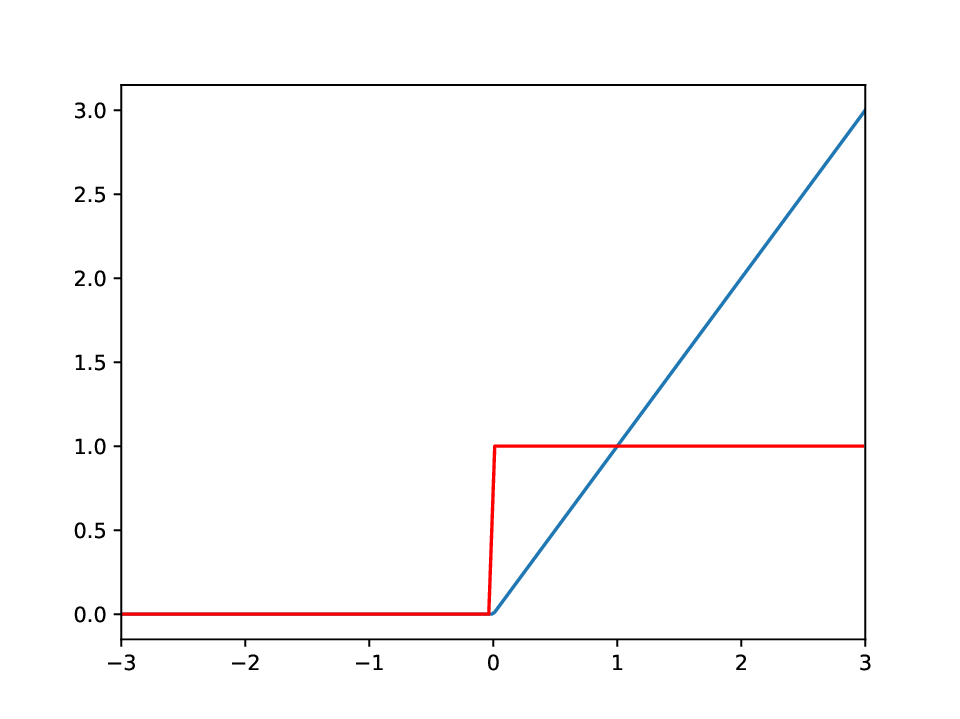}}
  \subfloat[Leaky-ReLU][]{\includegraphics[width=0.5\textwidth]{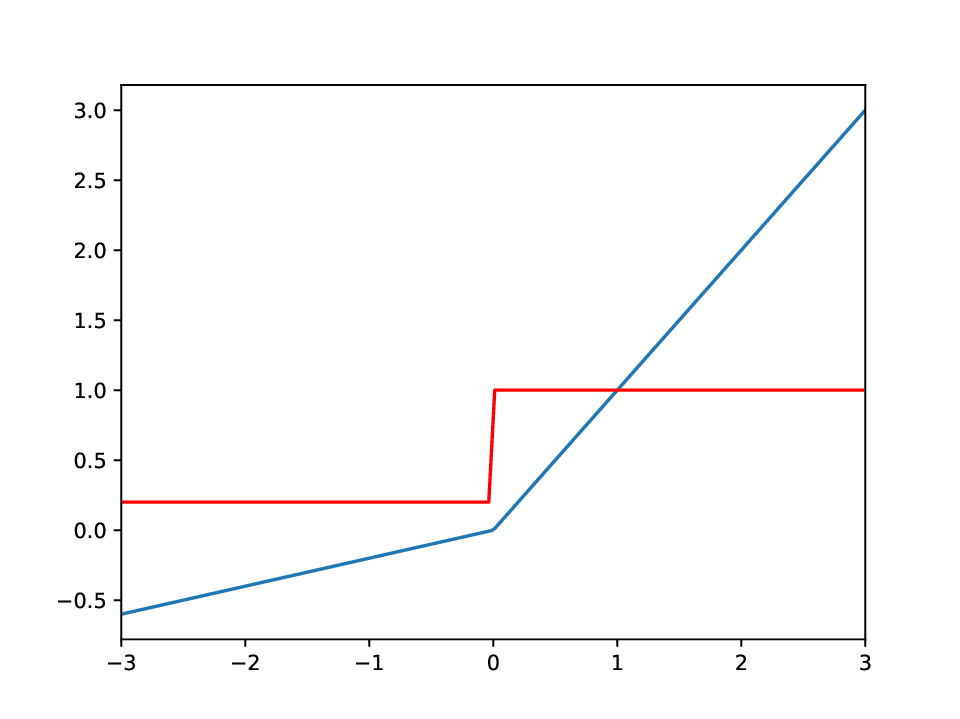}}
  \caption{Examples of commonly used activation functions in NNs. The
    output for each activation is shown in blue and the numerical
    derivative of each function is shown in red. These functions are (a) Sigmoid; (b) TanH; (c) ReLU; (d) Leaky-ReLU. Note the change in
    scale for the y-axis.}
  \label{fig:activation}
\end{figure}

Equation~\ref{eq:basis} refers to the state of each neuron (or node)
in the hidden layer. This is expressed as an affine transform followed
by a non-linear element wise transform $\phi(\cdot)$, which is often
called an activation. For the original perceptron, activation function
used was the $\text{sign}(\cdot)$ function, though the use of this
function has ceased due to it's derivative being equal to
zero\footnote{When the derivative is defined, as is a
  piece-wise non-differentiable function at the origin.}. More favourable activation functions such as the Sigmoid, Hyperbolic Tangent (TanH), Rectified Linear Unit (ReLU) and Leaky-ReLU have since replaced this the sign function \cite{glorot2011deep, maas2013rectifier}. Figure~\ref{fig:activation} illustrates these functions along with their corresponding derivatives. When using the Sigmoid function, expression~\ref{eq:basis} is
equivalent to logistic regression, meaning that the output of the
network becomes the sum of multiple logistic regression
models.
\par
For a regression model, the function applied to the output $g(\cdot)$
will be the identity function\footnote{Meaning no activation is used
  on the output layer, $g(x) = x$.}, and for binary classification
will be a Sigmoid.
\par
Equations~\ref{eq:basis} and~\ref{eq:mlp} can be efficiently
implemented using matrix representations, and is often represented as
such in machine learning literature. This is achieved by stacking the input vector in our data set as a column in $\mb{X}$. Forward propagation can then be performed as,
\begin{align}
  \label{eq:hidden_mat}
  \mb{\Phi}  &= a(\mb{X}^T \mb{W}^1),
  \\
  \label{eq:mlp_mat}
  \mb{F}  &= g(\mb{\Phi} \mb{W}^2).
\end{align}
Whilst this matrix notation is more concise, the choice to use the summation
notation to describe the network here is deliberate. It is hoped that with the summation notation, relations to kernel and statistical theory discussed later in this chapter becomes clearer.
\par
In the frequentist setting of NN learning, a MLE or MAP estimate is
found through the minimisation of a non-convex cost function
$J(x,y)$ w.r.t. network weights. Minimisation of this cost-function is performed through
backpropagation, where the output of the model is computed for the
current parameter settings, partial derivatives w.r.t parameters are
found and then used to update each parameter,
\begin{equation}
  \label{eq:backprop}
  w_{t+i} = w_{t} - \alpha \dfrac{\partial J(x,y)}{\partial w_t}.
\end{equation}
Equation~\ref{eq:backprop} illustrates how backpropagation updates
model parameters, with $\alpha$ representing the learning rate and the
subscripts indicate the iteration in the training procedure. Partial derivatives for individual parameters at different
layers in the network is found through application of the chain
rule. This leads to the preference of discontinuous non-linearities
such as the ReLU for deep NNs, as the larger gradient of the ReLU
assists in preventing vanishing gradients of early layers during
training.
\subsection{Bayesian Neural Networks}
In the frequentist setting presented above, the model weights are not treated as random variables; weights are assumed to have a true value that is just unknown and the data we have seen is treated as a random variable. This may seem conterintuitive for what we want to achieve. We would like to learn what our unknown model weights are based of the information we have at hand. For statistical modelling the information available to us comes in the form of our acquired data. Since we do not know the value for our weights, it seems natural to treat them as a random variable. The Bayesian view of statistics uses this approach; unknown (or latent) parameters are treated as random variables and we want to learn a distribution of these parameters conditional on the what we can observe in the training data.
\par
During the ``learning'' process of BNNs, unknown model weights are inferred based on what we do know or what we can observe.
This is the problem of inverse probability, and is solved through the use of Bayes Theorem.
The weights in our model $\bs{\omega}$ are hidden or latent variables; we cannot immediately observe their true distribution. Bayes Theorem allows us to represent a distribution over these weights in terms of probabilities we can observe, resulting in the distribution of model parameters conditional on the data we have seen $p(\bs{\omega} | \mathcal{D})$\footnote{$\mathcal{D}$ is used here to denote the set of training data $(\mb{x}, \mb{y})$.}, which we call the posterior distribution.
\par
Before training, we can observe the joint distribution between our weights and our data $p(\bs{\omega}, \mathcal{D})$. This joint distribution is defined by our prior beliefs over our latent variables $p(\bs{\omega})$ and our choice of model/likelihood $p(\mathcal{D} | \bs{\omega})$,
\begin{equation}
  \label{eq:joint}
  p(\bs{\omega}, \mathcal{D}) = p(\bs{\omega})p(\mathcal{D} | \bs{\omega}).
\end{equation}
Our choice of network architecture and loss function is used to define the likelihood term in Equation \ref{eq:joint}. For example, for a 1-D homoscedastic regression problem with a mean squared error loss and a known noise variance, the likelihood is a Gaussian distribution with the mean value specified by the output of the network,
\begin{equation*}
  p(\mathcal{D} | \bs{\omega}) = \mathcal{N}\big(\mathbf{f}^\omega(\mathcal{D}), \sigma^2\big).
\end{equation*}
Under this modelling scheme, it is typically assumed that all samples from $\mathcal{D}$ are i.i.d., meaning that the likelihood can then be written as a product of the contribution from the $N$ individual terms in the data set,
\begin{equation}
  \label{eq:like_prod}
  p(\mathcal{D} | \bs{\omega}) = \prod_{i=1}^{N}\mathcal{N}\big(\mathbf{f}^\omega(\mathbf{x}_i), \sigma^2\big).
\end{equation}
Our prior distribution should be specified to incorporate our belief as to how the weights should be distributed, prior to seeing any data. Due to the black-box nature of NNs, specifying a meaningful prior is challenging. In many practical NNs trained under the frequentist scheme, the weights of the trained network have a low magnitude, and are roughly centred around zero. Following this empirical observation, we may use a zero mean Gaussian with a small variance for our prior, or a spike-slab prior centred at zero to encourage sparsity in our model.
\par
With the prior and likelihood specified, Bayes theorem is then applied to yield the posterior distribution over the model weights,
\begin{equation}
  \label{eq:posterior}
  \pi(\bs{\omega} | \mathcal{D})
  = \dfrac{p(\bs{\omega})p(\mathcal{D} | \bs{\omega})}{\int p(\bs{\omega})p(\mathcal{D} | \bs{\omega}) d\bs{\omega}}
  = \dfrac{p(\bs{\omega})p(\mathcal{D} | \bs{\omega})}{p(\mathcal{D})}.
\end{equation}
The denominator in the posterior distribution is called the marginal likelihood, or the evidence. This quantity is a constant with respect to the unknown model weights, and normalises the posterior to ensure it is a valid distribution.
\par
From this posterior distribution, we can perform predictions of any quantity of interest. Predictions are in the form of an expectation with respect to the posterior distribution,
\begin{equation}
  \label{eq:expectations}
  \mathbb{E}_{\pi}[f] = \int f(\bs{\omega}) \pi(\bs{\omega} | \mathcal{D}) d\bs{\omega}.
\end{equation}
All predictive quantities of interest will be an expectation of this form. Whether it be a predictive mean, variance or interval, the predictive quantity will be an expectation over the posterior. The only change will be in the function $f(\bs{\omega})$ with which the expectation is applied to. Prediction can then be viewed as an average of the function $f$ weighted by the posterior $\pi(\bs{\omega})$.
\par
We see that the Bayesian inference process revolves around marginalisation (integration) over our unknown model weights. By using this marginalisation approach, we are able to learn about the generative process of a model, as opposed to an optimisation scheme used in the frequentist setting. With access to this generative model, our predictions are represented in the form of valid conditional probabilities.
\par
In this description, it was assumed that many parameters such as the noise variance $\sigma$ or any prior parameters were known. This is rarely the case, and as such we need to perform inference for these unknown variables. The Bayesian framework allows us to perform inference over these variables similarly to how we perform inference over our weights; we treat these additional variables as latent variables, assign a prior distribution (or sometimes called a hyper-prior) and then marginalise over them to find our posterior. For more of a description of how this can be performed for BNNs, please refer to \cite{lampinen2001bayesian, neal1996bayesian}.
\par
For many models of interest, computation of the posterior (Equation \ref{eq:posterior}) remains intractable. This is largely due to the computation of the marginal likelihood. For non-conjugate models or those that are non-linear in the latent variables (such as NNs), this quantity can be analytically intractable. For high dimensional models, a quadrature approximation of this integral can become computationally intractable. As a result, approximations for the posterior must be made. The following sections detail how approximate Bayesian inference can be achieved in BNNs.
\subsection{Origin of Bayesian Neural Networks}
~\label{sec:origin}
From this survey and those conducted prior \cite{gal2016uncertainty}, the first instance of what could be considered a
BNN was developed in \cite{tishby1989}. This paper emphasises key
statistical properties of NNs by developing a statistical
interpretation of loss functions used. It was shown that minimisation
of a squared error term is equivalent to finding the Maximum
Likelihood Estimate (MLE) of a Gaussian. More importantly, it was
shown that by specifying a prior over the network weights, Bayes Theorem
can be used to obtain an appropriate posterior. Whilst this
work provides key insights into the Bayesian perspective of NNs, no
means for finding the marginal likelihood (evidence) is supplied,
meaning that no practical means for inference is suggested. Denker and LeCun \cite{denker1991transforming} extend on this work,  offering a practical means for performing approximate inference using the Laplace
approximation, though minimal experimental results are provided.

\par
A NN is a generic function approximator. It is well known that as the
limit of the number of parameters approaches infinity in a single
hidden layer network, any arbitrary function can be represented
\cite{cybenko1989approximation, funahashi1989approximate,
  hornik1991approximation}. This means that for the practical case,
our finite training data set can be well approximated by a single
layer NN as long as there are sufficient trainable parameters in the
model. Similar to high-degree polynomial regression, although we can
represent any function and even exactly match the training data in
certain cases, as the number of parameters in a NN increases or the degree of the polynomial used increases, the
model complexity increases leading to issues of overfitting. This
leads to a fundamental challenge found in NN design; how complex
should I make my model?
\par
Building on the work of Gull and Skilling \cite{gull1999quantified},
MacKay demonstrates how a Bayesian framework naturally lends itself to
handle the task of model design and comparison of generic statistical
models \cite{mackay1992interp}. In this work, two levels of inference
are described: inference for fitting a model and inference for
assessing the suitability of a model. The first level of inference is
the typical application of Bayes rule for updating model parameters,
\begin{equation}
  \label{eq:mackayBayes}
  P(\bs{\omega} | \mathcal{D}, \mathcal{H}_i) = \dfrac{P(\mathcal{D} | \bs{\omega}, \mathcal{H}_i) P(\bs{\omega} | \mathcal{H}_i)}{ P(\mathcal{D} | \mathcal{H}_i)},
\end{equation}
where $\bs{\omega}$ is the set of parameters in the generic statistical
model, $\mathcal{D}$ is our data and $\mathcal{H}_i$ represents the
$i$'th model used for this level of inference\footnote{$\mathcal{H}$
  is used to refer to the model ``hypothesis''.}. This is then
described as,
\begin{equation}\nonumber
  \label{eq:mackayEvidence}
  \text{Posterior} = \dfrac{\text{Likelihood} \times \text{Prior}}{\text{Evidence}}.
\end{equation}
It is important to note that the normalising constant in Equation~\ref{eq:mackayBayes}
is referred to as the evidence for the specific
model of interest $\mathcal{H}_i$. Evaluation of the posterior remains
intractable for most models of interest, so approximations must be
made. In this work, the Laplace approximation is used.
\par
\begin{figure}[!h]
  \centering
  \includegraphics[width=0.6\linewidth]{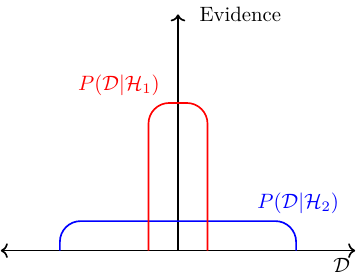}
  \caption{Graphical illustration of how the evidence plays a role in
    investigating different model hypotheses. The simple model
    $\mathcal{H}_1$ is able to predict a small range of data with
    greater strength, while the more complex model $\mathcal{H}_2$ is
    able to represent a larger range of data, though with lower
    probability. Adapted from \cite{mackay1992interp, mackay1992bayesian}.}
  \label{fig:evidence}
\end{figure}

Though computation of the posterior over parameters is required, the
key aim of this work is to demonstrate methods of assessing the
posterior over the model hypothesis $\mathcal{H}_i$. The posterior
over model design is represented as,
\begin{equation}
  \label{eq:mackayHypothesis}
  P(\mathcal{H}_i | \mathcal{D}) \propto P(\mathcal{D} | \mathcal{H}_i)  P(\mathcal{H}_i),
\end{equation}
which translates to,
\begin{equation}\nonumber
  \label{eq:mackayHypothesisText}
  \text{Model Posterior} \propto \text{Evidence} \times \text{Model Prior}.
\end{equation}
The data dependent term in Equation~\ref{eq:mackayHypothesis} is the evidence
for the model. Despite the promising interpretation of the posterior
normalisation constant, as described earlier, evaluation of this distribution is intractable for most BNNs. Assuming a Gaussian distribution, the
Laplace approximation of the evidence can be found as,
\begin{align}
  \label{eq:mackayEvidenceApprox}
  P(\mathcal{D} | \mathcal{H}_i)
  & = \int P(\mathcal{D} | \bs{\omega}, \mathcal{H}_i) P(\bs{\omega} | \mathcal{H}_i) d\bs{\omega}
  \\
  & \approx P (\mathcal{D} | \bs{\omega}_{\text{MAP}}, \mathcal{H}_i) \Big[ P(\bs{\omega}_{\text{MAP}} | \mathcal{H}_i) \Delta\bs{\omega}\Big]
  \\
  & = P (\mathcal{D} | \bs{\omega}_{\text{MAP}}, \mathcal{H}_i) \Big[ P(\bs{\omega}_{\text{MAP}} | \mathcal{H}_i) (2 \pi)^{\frac{k}{2}} \text{det}^{-\frac{1}{2}}\mathbf{A} \Big]
  \\
  & = \text{Best Likelihood Fit} \times \text{Occam Factor}. \nonumber
\end{align}
This can be interpreted as a single Riemann approximation to the model
evidence with the best likelihood fit representing the peak of the
evidence, and the Occam factor is the width that is characterised by
the curvature around the peak of the Gaussian. The Occam factor can be
interpreted as the ratio of the width of the posterior
$\Delta \bs{\omega}$ and the range of the prior $\Delta \bs{\omega}_0$
for the given model $\mathcal{H}_i$,
\begin{equation}
  \label{eq:occam}
  \text{Occam Factor} = \dfrac{\Delta \bs{\omega}}{\Delta \bs{\omega}_0},
\end{equation}
meaning that the Occam factor is the ratio of change in plausible
parameter space from the prior to the posterior. Figure \ref{fig:evidence} demonstrates this concept graphically. With this
representation, a complex model able to represent a large range of
data will have a wider evidence, thus having a larger Occam factor. A
simple model will have a lower capacity to capture a complex
generative process, but a smaller range of data will be able to be
modelled with greater certainty, resulting in a lower Occam
Factor. This results in a natural regularisation for the complexity of
a model. An unnecessarily complex model will typically result in a
wide posterior, resulting in a large Occam factor and low evidence
for the given model. Similarly, a wide or less informative prior will result in a
reduced Occam factor, providing further intuition into the Bayesian
setting of regularisation.
\par
Using this evidence framework requires computation of the marginal likelihood, which is an expensive (and the key challenge) within Bayesian modelling. Given the large investment required to approximate the marginal likelihood, it may be infeasible to compare many different architectures. Despite this, the use of the evidence framework can used to assess solutions for BNNs. For most NN architectures of interest,
the objective function is non-convex with many local minima. Each local minima can be regarded as a possible solution for the inference problem. MacKay uses this as motivation to compare the solutions from each local minimum using the corresponding evidence function \cite{mackay1992practical}. This allows for assessment of model complexity at each solution without prohibitive computational requirements.
\par
\subsubsection{Early Variational Inference for BNNs}
The machine learning community has continuously excelled at
optimisation based problems. While many ML models, such as Support
Vector Machines and Linear Gaussian Models result in a convex
objective function, NNs have a highly non-convex objective function
with many local minima. A difficult to locate global minimum motivates
the use of gradient based optimisation schemes such as
backpropagation \cite{rumelhart1986learning}. This type of optimisation can be viewed in a Bayesian
context through the lens of Variational Inference (VI).
\par
VI is an approximate inference method that frames marginalisation required during Bayesian inference as an optimisation problem
\cite{jordan1999introduction, wainwright2008graphical,
  blei2017variational}. This is achieved by assuming the form of the posterior
distribution and performing optimisation to find the assumed density that closest to the true posterior. This assumption simplifies computation and provides some level of tractability.
\par
The assumed posterior distribution $q_{\bs{\theta}}(\bs{\omega})$ is a suitable density over the set of parameters $\mb{\omega}$, that is
restricted to a certain family of distributions parameterised by
$\mathbf{\theta}$. The parameters for this variational distribution
are then adjusted to reduce the dissimilarity between the variational
distribution and the true posterior
$p(\bs{\omega} | \mathcal{D} )$\footnote{The model hypothesis
  $\mathcal{H}_i$ used previously will be omitted for further
  expressions, as little of the remaining key research items deal with
  model comparison and simply assume a single architecture and
  solution.}. The means to measure similarity for VI is
often the forward KL-Divergence between the variational and true
distribution,
\begin{equation}
  \label{eq:kl}
  KL\Big(q_{\bs{\theta}}(\bs{\omega}) || p(\bs{\omega} | \mathcal{D}) \Big) = \int q_{\bs{\theta}}(\bs{\omega}) \log \dfrac{q_{\bs{\theta}}(\bs{\omega})}{p(\bs{\omega} | \mathcal{D})} d\bs{\omega}.
\end{equation}
For VI, Equation \ref{eq:kl} serves as the objective function we wish
to minimise w.r.t variational parameters $\mathbf{\theta}$. This can
be expanded out as,
\begin{align}
  \label{eq:KLRe}
  \text{KL}\Big( q_{\bs{\theta}}(\bs{\omega}) || p(\bs{\omega}|\mathcal{D}) \Big) =
  \ & {} \mathbb{E}_{q} \big[
      \log \dfrac{q_{\bs{\theta}}(\bs{\omega})}{p(\bs{\omega})} - \log p(\mathcal{D}|\bs{\omega})
      \big] + \log p(\mathcal{D})
  \\
  = & \ {} \text{KL}\Big( q_{\bs{\theta}}(\bs{\omega}) || p(\bs{\omega}) \Big) -
      \mathbb{E}_{q} [\log p(\mathcal{D}|\bs{\omega}) ] + \log p(\mathcal{D})
  \\
  \label{eq:KLF}
  = & \ - \mathcal{F}[q_{\mathbf{\theta}}] + \log p(\mathcal{D}),
\end{align}
where
$\mathcal{F}[q_{\mathbf{\theta}}] = - \text{KL}\Big(
q_{\bs{\theta}}(\bs{\omega}) || p(\bs{\omega}) \Big) +
\mathbb{E}_{q} [\log p(\mathcal{D}|\bs{\omega})]$. The combination of terms into
$\mathcal{F}[q]$ is to separate the tractable terms from the
intractable log marginal likelihood. We can now optimise this function
using backpropagation, and since the log marginal likelihood does not
depend on variational parameters $\mathbf{\theta}$, it's derivative evaluates to
zero. This leaves only term of containing variational parameters, which is $\mathcal{F}[q_{\mathbf{\theta}}]$.
\par
\begin{figure}[!h]
  \centering \includegraphics[width=0.6\linewidth]{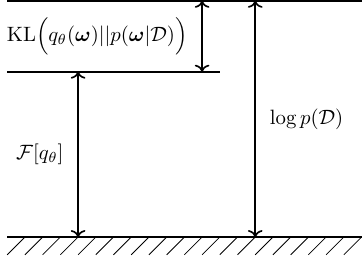}
  \caption{Graphical illustration of how the minimisation of the KL
    divergence between the approximate and true posterior maximises
    the lower bound on the evidence. As the KL Divergence between our
    approximate and true posterior is minimised, the ELBO
    $\mathcal{F}[q_{\mathbf{\theta}}]$ tightens to the
    log-evidence. Therefore maximising the ELBO is equivalent to
    minimising the KL divergence between the approximate and true
    posterior. Adapted from \cite{barber1998ensemble}.}
  \label{fig:elbo}
\end{figure}

This notation used in Equation \ref{eq:KLF}, particularly the choice to include
the negative of $\mathcal{F}[q_{\mathbf{\theta}}]$ is deliberate to
highlight a different but equivalent derivation to the identical
result, and to remain consistent with existing literature. This result can be obtained by instead of
minimising the KL-Divergence between the true and approximate distribution, but by approximating the intractable
log marginal likelihood. Through application of Jensen's inequality,
we can then find that $\mathcal{F}[q_{\mathbf{\theta}}]$ forms a lower
bound on the logarithm of the marginal likelihood \cite{jordan1999introduction, hoffman2013stochastic}. This can be seen by
re-arranging Equation \ref{eq:KLF} and noting that the KL divergence is strictly $\geq 0$ and only equals zero when the two distributions are equal. The logarithm of the marginal
likelihood is equal to the sum of the KL divergence between the
approximate and true posterior and
$\mathcal{F}[q_{\mathbf{\theta}}]$. By minimising the KL divergence
between the approximate and true posterior, the closer
$\mathcal{F}[q_{\mathbf{\theta}}]$ will be to the logarithm of the
marginal likelihood. For this reason,
$\mathcal{F}[q_{\mathbf{\theta}}]$ is commonly referred to as the
Evidence Lower Bound (ELBO). Figure~\ref{fig:elbo} illustrates this graphically.
\par
The first application of VI to BNNs was by Hinton and Van Camp
\cite{hinton1993keeping}, where they tried to address the problem of
overfitting in NNs. They argued that by using a probabilistic
perspective of model weights, the amount of information they could
contain would be reduced and would simplify the network. Formulation
of this problem was through an information theoretic basis,
particularly the Minimum Descriptive Length (MDL) principle, though its
application results in a framework equivalent to VI. As is common in
VI, the mean-field approach was used. Mean-Field Variational Bayes
(MFVB) assumes a posterior distribution that factorises over
parameters of interest. For the work in \cite{hinton1993keeping}, the
posterior distribution over model weights was assumed to be a
factorisation of independent Gaussians,
\begin{equation}
  \label{eq:factor}
  q_{\bs{\theta}}(\bs{\omega}) = \prod_{i=1}^{P} \mathcal{N}(w_i | \mu_i, \sigma_i^2),
\end{equation}
where $P$ is the number of weights in the network.
For a regression network with a single hidden layer, an analytic
solution for this posterior is available. The ability to achieve an analytic solution to the approximation is an desirable property, as analytic solutions significantly reduce the time to perform inference.
\par
There are a few issues with this work, though one of the most
prominent issues is the assumption of a posterior that factorises over
individual network weights. It is well known that strong correlation
between parameters in a NN is present. A factorised distribution
simplifies computation by sacrificing the rich correlation information
between parameters. MacKay highlighted this limitation in an early
survey of BNNs \cite{mackay1995probable} and offers insight into how a
preprocessing stage of inputs to hidden layers could allow for more
comprehensive approximate posterior distributions.
\par
Barber and Bishop \cite{barber1998ensemble} again highlight this
limitation, and offer a VI based approach that extends on the work in
\cite{hinton1993keeping} to allow for full correlation between
the parameters to be captured by using a full rank Gaussian for the
approximating posterior. For a single hidden layer regression network
utilising a Sigmoid activation, analytic expressions for evaluating the
ELBO is provided\footnote{Numerical methods are required to evaluate
  certain terms in the analytic expression for the ELBO.}. This is
achieved by replacing the Sigmoid with the appropriately scaled error
function.
\par
An issue with this modelling scheme is the increased number
of parameters. For a full covariance model, the number of parameters
scales quadratically with the number of weights in the network. To
rectify this, Barber and Bishop propose a restricted form for the
covariance often used in factor analysis, such that,
\begin{equation}
  \label{eq:restCov}
  \mathbf{C} = \text{diag}(d_1^2, ..., d_n^2) + \sum_{i = 1}^s \mathbf{s}_i\mathbf{s}_i^T,
\end{equation}
where the diag operator creates a diagonal matrix from the vector
$\mathbf{d}$ of size $n$, where $n$ is the number of weights in the
model. This form then scales linearly with the number of hidden units
in the network.
\par
These bodies of work provide important insight into how the prominent
backpropagation method can be applied to challenging Bayesian
problems. This allows for properties of the two areas of research to
be merged and offer the benefits nominally seen in isolation. Complex
regression tasks for large bodies of data sets could now be handled
in a probabilistic sense using NNs.
\par
Despite the insight offered by these methods, there are
limitations to these methods. Both the work of Hinton and Van Camp and
Barber and Bishop focus on development of a closed form representation
of the networks\footnote{Although there are a large number of benefits
  to such an approach, as illustrated earlier.}. This analytic tractability imposes many restrictions on the networks. As
discussed previously, \cite{hinton1993keeping} assume a factorised
posterior over individual weights which is unable to capture any
correlation in parameters. Covariance structure is captured in \cite{barber1998ensemble}, though the authors limit their analysis to the use of a Sigmoid activation function (which is well
approximated by the error function), which is seldom used in modern
networks due to the low magnitude in the
gradient\footnote{Analytic results may be achievable using other
  activation functions, such as TanH, which suffer less from such an
  issue.}. A key limitation common to both of these approaches is the
restriction of a single hidden layer network.
\par
As stated previously, a NN can approximate any function arbitrarily
well by adding additional hidden units. For modern networks, empirical
results have shown that similarly complex functions can be represented
with fewer hidden units by increasing the number of hidden layers in
the network. This has lead to the term ``deep learning'', where depth
refers to the number of hidden layers. The reduction in number of weight variables is
especially important for when trying to approximate the full covariance
structure between layers. For example, correlation between hidden
units within a single layer may be captured, while assuming that
parameters between the different layers are independent. An assumption
such as this can significantly reduce the number of correlation
parameters. With modern networks having hundreds of millions of
weights across many layers (with these networks only being able to
offer point estimates), the need to develop practical probabilistic
interpretations beyond a single layer is essential.
\subsubsection{Hybrid Monte Carlo for BNNs}
It is worthwhile at this point to reflect on the actual quantities of interest. So far the emphasis has been placed on finding good approximations for the posterior, though the accurate representation of the posterior is usually not the end design requirement. The main quantities of interest are predictive moments and intervals. We want to make good predictions accompanied by confidence information. The reason we emphasise computation of the  posterior is that predictive moments and intervals are all computed as expectations of the posterior $\pi(\bs{\omega}|\mathcal{D})$\footnote{Note that $\pi$ is used to represent the true posterior distribution here, as appose to $q$ used previously to denote an approximation of the posterior.}. This expectation is listed in Equation \ref{eq:expectations}, and is repeated here for convenience,
\begin{equation*}
  \mathbb{E}_{\pi}[f] = \int f(\bs{\omega}) \pi(\bs{\omega} | \mathcal{D}) d\bs{\omega}.
\end{equation*}
This is why computation of the posterior is emphasised; accurate predictions rely on accurate approximations of the intractable posterior.
\par
The previous methods employed optimisation based schemes such as VI or Laplace approximations of the posterior. In doing so, strong assumptions and restrictions on the form of posterior are enforced. The restrictions placed are often credited with inaccuracies induced in predictions, though this is not the only limitation.
\par
As highlighted by~\cite{betancourt2017conceptual, betancourt2017geometric}, the expectation computed for predictive quantities not just a probability mass, it the product of the probability mass and a volume. The probability mass is our posterior distribution $\pi(\bs{\omega} | \mathcal{D})$, and the volume $d\bs{\omega}$ over which we are integrating. It is likely that for all models of interest, the contribution of the expectation from this product of the density and volume will not be at the maximum for the mass. Therefore optimisation based schemes which consider only the mass can deliver inaccurate predictive quantities. To make accurate predictions with finite computational resources, we need to evaluate this expectation not just when the mass is greatest, but when the product of the mass and volume is largest. The most promising way to achieve this is with Markov Chain Monte Carlo (MCMC).
\par
MCMC algorithms remains at the forefront of Bayesian
research and applied statistics\footnote{MCMC is regarded as one of the most influential algorithms of the 21st century \cite{madey2005agent}.}. MCMC is a general approach for sampling from arbitrary and intractable distributions. The ability to sample from a distribution enables the use of Monte Carlo integration for prediction,
\begin{equation}
  \label{eq:expectations_mc}
  \mathbb{E}_{\pi}[f] = \int f(\bs{\omega}) \pi(\bs{\omega} | \mathcal{D}) d\bs{\omega} \approx
  \dfrac{1}{N} \sum_{i = 1}^{N} f(\bs{\omega}_i),
\end{equation}
where $\bs{\omega}_i$ represents an independent sample from the posterior distribution. MCMC enables sampling from our posterior distribution, with the samples converging to when the product of the probability density and volume are greatest~\cite{betancourt2017conceptual}.
\par
Assumptions previously made in VI methods,
such as a factorised posterior are not required in the MCMC
context. MCMC provides convergence to the true posterior as the
number of samples approaches infinity. By avoiding such restrictions, with enough time and computing resources we can yield a solution that is closer to the true predictive quantities. This is an important challenge for BNNs, as the
posterior distributions is typically quite complex.
\par
Traditional MCMC methods demonstrate a random-walk behaviour, in that new proposals in the sequence are generated randomly.
Due to the complexity and high dimension of the posterior in BNNs, this random-walk behaviour makes these methods
unsuitable for performing inference in any reasonable time.
To avoid the random-walk behaviour, Hybrid/Hamiltonian Monte Carlo
(HMC) can be employed to incorporate gradient information into the
iterative behaviour. While HMC was initially proposed for statistical
physics~\cite{duane1987hybrid}, Neal highlighted the potential
for HMC to address Bayesian inference and specifically researched the
applications to BNNs and the wider statistics community as a whole~\cite{neal1996bayesian}.
\par
Given that HMC was initially proposed for physical dynamics, it is
appropriate to build intuition for applied statistics through a
physical analogy. Treat our parameters of interest
$\bs{\omega}$ as a position variable. An auxiliary variable is then
introduced to model the momentum $\mb{v}$ of our current
position. This auxiliary variable is not of statistical interest, and is only introduced to aid in development of the system dynamics.
With a position and momentum variable, we can represent the
potential energy $U(\bs{\omega})$ and the kinetic energy $K(\mb{v})$ of
our system. The total energy of a system is then represented as,
\begin{equation}
  \label{eq:energy}
  H(\bs{\omega}, \mb{v}) = U(\bs{\omega}) + K(\mb{v}).
\end{equation}
We now consider the case of a lossless system, in that the total
energy $H(\bs{\omega}, \mb{v})$ is constant\footnote{The values for
  $\bs{\omega}$ and $\mb{v}$ will change, though the total energy of the
  system will remain constant}. This is described as a Hamiltonian
system, and is represented as the following system of differential
equations \cite{neal2011mcmc},
\begin{align}
  \label{eq:hmc}
  \dfrac{dw_i}{dt} &= \dfrac{\partial H}{\partial v_i},
  \\
  \dfrac{dv_i}{dt} &= - \dfrac{\partial H}{\partial w_i},
\end{align}
where $t$ represents time and the $i$ denotes the individual
elements in $\bs{\omega}$ and $\mb{v}$.
\par
With the dynamics of the system defined, we wish to relate the physical interpretation to a probabilistic interpretation. This can be
achieved through the canonical distribution\footnote{As is commonly done, we assume the
  temperature variable included in physical representations of the
  canonical distribution is set to one. For more information, see \cite[p. 11]{neal2011mcmc}, \cite[p. 123]{brooks2011handbook}.},
\begin{equation}
  \label{eq:cononical}
  P(\bs{\omega}, \mb{v}) = \dfrac{1}{Z}\exp\big(- H(\bs{\omega}, \mb{v})\big) = \dfrac{1}{Z}\exp\big(- U(\bs{\omega})\big)\exp\big( -K(\mb{v})\big),
\end{equation}
where $Z$ is a normalising constant and $H(\bs{\omega}, \mb{v})$ is our
total energy as defined in Equation~\ref{eq:energy}. From this joint
distribution, we see that our position and momentum variable are
independent.
\par
Our end goal is to find predictive moments and intervals. For a Bayesian this makes the key quantity of interest the
posterior distribution. Therefore, we can set the potential energy
which we wish to sample from to,
\begin{equation}
  \label{eq:posterior_potential}
  U(\bs{\omega}) = -\log\Big(p(\bs{\omega})p(\mathcal{D} | \bs{\omega}) \Big).
\end{equation}
Within HMC, the kinetic energy can be freely selected from a wide range of suitable functions, though is typically
chosen such that it's marginal distribution of $\mb{v}$ is a diagonal
Gaussian centred at the origin.
\begin{equation}
  \label{eq:kinetic}
  K(\mb{v}) = \mb{v}^T M^{-1} \mb{v},
\end{equation}
where $M$ is a diagonal matrix referring to the ``mass'' of our
variables in this physical interpretation. Although this is the most common kinetic energy function used, it may not be the most suitable. \cite{betancourt2017conceptual} surveys the selection the design of other Gaussian kinetic energies with an emphasis on the geometric interpretations. It is also highlighted that selection of appropriate kinetic energy functions remains an open research topic, particularly in the case of non-Gaussian functions.
\par
Since Hamiltonian dynamics leaves the total energy invariant, when
implemented with infinite precision, the dynamics proposed are
reversible. Reversibility is a sufficient property to satisfy the
condition of detailed balance, which is required to ensure that the
target distribution (the posterior we are trying to sample from)
remains invariant. For practical implementations, numerical errors
arise due to discretisation of variables. The discretisation method
most commonly employed is the leapfrog method. The leapfrog method
specifies a step size $\epsilon$ and a number of steps $L$ to be used
before possibly accepting the new update. The leapfrog method first
performs a half update of the momentum variable $v$, followed by a
full update of the position $w$ and then the remaining half update of
the momentum \cite{neal2011mcmc},
\begin{align}
  v_i(t + \frac{\epsilon}{2}) = v_i(t) + \frac{\epsilon}{2} \dfrac{dv_{i}(t)}{dt},
  \\
  w_i(t + \epsilon) = w_i(t) + \epsilon \frac{dw_{i}(t + \epsilon /2)}{dt},
  \\
  v_i(t + \epsilon) = v_i(t +\frac{\epsilon}{2}) + \frac{\epsilon}{2} \dfrac{dv_{i}(t + \epsilon)}{dt}.
\end{align}
If the value of $\epsilon$ is chosen such that this dynamical system
remains stable, it can be shown that this leapfrog method preserves
the volume (total energy) of the Hamiltonian.
\par
For expectations to be approximated using~\ref{eq:expectations_mc}, we require each sample $\bs{\omega}_i$ to be independent from subsequent samples. We can achieve practical independence\footnote{Where for all practical purposes each sample can be viewed as independent.} by using multiple leapfrog steps $L$. In this way, after $L$ leapfrog steps of size $\epsilon$, the new position is proposed. This reduces correlation between samples and can allow for faster exploration of the posterior space. A Metropolis step is then applied to determine whether
this new proposal is accepted as the newest state in the Markov
Chain \cite{neal2011mcmc}.
\par
For the BNN proposed by \cite{neal1996bayesian}, a hyper-prior $p(\bs{\gamma})$ is induced to model the variance over prior parameter precision and likelihood precision. A Gaussian prior is used for the prior over-parameters and the likelihood is set to be Gaussian. Therefore, the prior over the $\bs{\gamma}$ was Gamma distributed, such that it was conditionally conjugate. This allows for Gibbs sampling to be used for performing inference over hyperparameters. HMC is then used to update the posterior parameters. Sampling from the joint posterior $P(\bs{\omega}, \bs{\gamma} | \mathcal{D})$ then involves alternating between the Gibbs sampling step for the hyperparameters and Hamiltonian dynamics for the model parameters. Superior performance of HMC for simple BNN models was then demonstrated and compared with random walk MCMC and Langevin methods \cite{neal1996bayesian}.
\subsection{Modern BNNs}
Considerably less research was conducted into BNNs following  early work of Neal, MacKay and Bishop proposed in the 90s. This relative
stagnation was seen throughout the majority of NN research, and was
largely due to the high computational demand for training NNs. NNs are
parametric models that are able to capture any function with arbitrary
accuracy, but to capture complex functions accurately requires large
networks with many parameters. Training of such large networks became
infeasible even for the traditional frequentist perspective, and the
computational demand significantly increases to investigate the more
informative Bayesian counterpart.
\par
Once it was shown that general purpose GPUs could accelerate and allow
training of large models, interest and research into NNs saw a
resurgence. GPUs enabled large scale parallelism of the linear algebra
performed during back propagation. This accelerated computation has
allowed for training of deeper networks, where successive concatenation
of hidden layers is used. With the proficiency of GPUs for optimising
complex networks and the great empirical success seen by such models,
interest into BNNs resumed.
\par
Modern research into BNNs has largely focused on the VI approach,
given that these problems can be optimised using a similar
backpropagation approach used for point estimate networks. Given that
the networks offering the most promising results use multiple layers,
the original VI approaches shown in \cite{hinton1993keeping,
  barber1998ensemble}, which focus on analytical approximations for
regression networks utilising a single hidden layer became unsuitable. Modern NNs now
exhibit considerably different architectures with varying
dimensions, hidden layers, activations and applications. More general
approaches for viewing networks in a probabilistic sense was required.
\par
Given the large scale of modern networks, large data sets are
typically required for robust inference\footnote{Neal \cite{neal1996bayesian} argues that this not true for Bayesian modelling; claims that if suitable prior information is available, complexity of a model should only be limited by computational resources.}. For these large data sets,
evaluation of the complete log-likelihood becomes infeasible for
training purposes. To combat this, a Stochastic Gradient Descent (SGD)
approach is used, where mini-batches of the data are used to
approximate the likelihood term, such that our variational objective
becomes,

\begin{equation}
  \label{eq:vi_sgd}
  \mathcal{L}(\bs{\omega}, \bs{\theta}) = -\dfrac{N}{M} \sum_{i=1}^N
  \mathbb{E}_{q} [\log\big( p(\mathcal{D}_i|\bs{\omega}) \big)] +
  \text{KL}\Big( q_{\bs{\theta}}(\bs{\omega}) || p(\bs{\omega}) \Big),
\end{equation}
where $\mathcal{D}_i \subset \mathcal{D}$, and each subset is of size
$M$. This provides an efficient way to utilise large data sets during
training. After passing a single subset $\mathcal{D}_i$,
backpropagation is applied to update the model parameters. This
sub-sampling of the likelihood induces noise into our inference
process, hence the name SGD. This noise that is induced is expected to
average out over evaluation of each individual subset \cite{welling2011sgld}. SGD is the most common method for training NNs and
BNNs utilising a VI approach.
\par
A key paper in the resurgence of BNN research was published by
Graves \cite{graves2011}. This work proposes a MFVB treatment using a
factorised Gaussian approximate posterior. The key contribution of
this work is the computation of the derivatives. The VI objective
(ELBO) can be viewed as a sum of two expectations,
\begin{equation}
  \label{eq:elbo_expectation}
  \mathcal{F}[q_{\mathbf{\theta}}] =
  \mathbb{E}_{q} [\log\big( p(\mathcal{D}|\bs{\omega}) \big)]
  -\mathbb{E}_{q}[ \log q_{\bs{\theta}}(\bs{\omega}) -\log p(\bs{\omega})]
\end{equation}
It is these two expectations that we need to optimise w.r.t model
parameters, meaning that we require the gradient of expectations. This
work shows how using the gradient properties of a Gaussian proposed in
\cite{opper2009variational} can be used to perform parameter updates,
\begin{align}
  \label{eq:grad_expectation_mean}
  \nabla_{\bs{\mu}}\  \mathbb{E}_{p(\bs{\omega})}[f(\bs{\omega})]
  &= \mathbb{E}_{p(\bs{\omega})}[\nabla_{\bs{\omega}}f(\bs{\omega})],
  \\
  \label{eq:grad_expectation_var}
  \nabla_{\Sigma} \ \mathbb{E}_{p(\bs{\omega})}[f(\bs{\omega})]
  &= \dfrac{1}{2}\mathbb{E}_{p(\bs{\omega})}[\nabla_{\bs{\omega}} \nabla_{\bs{\omega}}f(\bs{\omega})].
\end{align}
MC integration could be applied to Equations
\ref{eq:grad_expectation_mean} and \ref{eq:grad_expectation_var} to
approximate the gradient of the mean and variance parameters. This framework allows for optimisation of the ELBO to generalise to any log-loss parametric model.
\par
Whilst addressing the problem of applying VI to complex BNNs with more
hidden layers, practical implementations have shown inadequate
performance which is attributed to large variance in the MC
approximations of the gradient computations
\cite{hernandez2015probabilistic}. Developing gradient estimates with
reduced variance has become an integral research topic in VI
\cite{paisley2012variational}. Two of the most common methods for
deriving gradient approximations rely on the use of score functions
and path-wise derivative estimators.
\par
Score function estimators rely on the use of the log-derivative
property, such that,
\begin{equation}
  \label{eq:logDiv}
  \dfrac{\partial}{\partial \theta} p(x | \theta) = p(x | \theta)   \dfrac{\partial}{\partial \theta} \log p(x | \theta).
\end{equation}
Using this property, we can form Monte Carlo estimates of the
derivatives of an expectation, which is often required in VI,
\begin{align}
  \nabla_\theta \mathbb{E}_{q} [ f(\omega) ]
  &= \int f(\omega) \nabla_\theta q_{{\theta}}(\omega) \partial\omega\nonumber
  \\
  &= \int f(\omega) q_{{\theta}}(\omega) \nabla_\theta \log\Big(q_{{\theta}}(\omega) \Big) \partial\omega\nonumber
  \\
  \label{eq:score}
  &\approx \dfrac{1}{L} \sum_{i=1}^L f(\omega_i) \nabla_\theta \log\Big(q_{{\theta}}(\omega_i) \Big).
\end{align}
A common problem with score function gradient estimators is that they
exhibit considerable variance \cite{paisley2012variational}. One of
the most common methods to reduce the variance in Monte Carlo
estimates is the introduction of control variates \cite{wilson1984}.
\par
The second type of gradient estimator commonly used in the VI
literature is the pathwise derivative estimator. This work builds on
the ``reparameterisation trick'' \cite{opper2009TheVG, kingma2013auto,
  rezende2014stochastic}, where a random variable is represented as a
deterministic and differentiable expression. For example, for a
Gaussian with $\bs{\theta} = \{ \mb{\mu}, \bs{\sigma} \}$,
\begin{align}
  \bs{\omega} &\sim \N(\bs{\mu}, \bs{\sigma}^2) \nonumber
  \\
  \bs{\omega} = g(\bs{\theta}, \bs{\epsilon}) &= \bs{\mu} + \bs{\sigma} \odot \bs{\epsilon}
                                                \label{eq:reparam_example}
\end{align}
where $\bs{\epsilon} \sim \N(\mb{0}, \mb{I})$ and $\odot$ represents
the Hadamard product.  Using this method allows for efficient sampling
for Monte Carlo estimates of expectations. This is shown in
\cite{kingma2013auto}, that with
$ \bs{\omega} = g(\bs{\theta}, \bs{\epsilon}) $, we know that
$ q(\bs{\omega} | \bs{\theta}) d\bs{\omega} =
p(\bs{\epsilon})d\bs{\epsilon}$. Therefore, we can show that,
\begin{align}
  \int q_{\bs{\theta}}(\bs{\omega}) f(\bs{\omega}) d\bs{\omega}
  &= \int p(\bs{\epsilon})  f(\bs{\omega}) d\bs{\epsilon}\nonumber
  \\
  &=  \int p(\bs{\epsilon}) f( g(\bs{\theta}, \bs{\epsilon})) d\bs{\epsilon}\nonumber
  \\
  \label{eq:MCtrick}
  \approx \dfrac{1}{M} \sum_{i=1}^M  f( g(\bs{\theta}, \bs{\epsilon}_i))
  &=  \dfrac{1}{M} \sum_{i=1}^M  f( \bs{\mu} + \bs{\sigma} \odot \bs{\epsilon}_i)
\end{align}
Since Equation~\ref{eq:MCtrick} is differentiable w.r.t
$\bs{\theta}$, gradient descent methods can be used to optimise this
expectation approximation. This is an important property in VI, since
the VI objective contains expectations of the log-likelihood that
are often intractable. The reparameterisation trick serves as the
basis for pathwise-gradient estimators. Pathwise estimators are
favourable for their reduced variance over score function estimators
\cite{kingma2013auto, paisley2012variational}.
\par
A key benefit of having a Bayesian treatment of NNs is the ability to
extract uncertainty in our models and their predictions. This has been
a recent research topic of high interest in the context of
NNs. Promising developments regarding uncertainty estimation in NNs
has been found by relating existing regularisation techniques such as
Dropout \cite{srivastava2014dropout} to approximate inference. Dropout
is a Stochastic Regularisation Technique (SRT) that was proposed to
address overfitting commonly seen in point-estimate networks. During
training, Dropout introduces an independent random variable that is
Bernoulli distributed, and multiplies each individual weight
element-wise by a sample from this distribution. For example, a simple
MLP implementing Dropout is of the form,
\begin{align}
  \rho_u & \sim \text{Bernoulli}(p), \nonumber
  \\
  \phi_j &=  \theta\Big(\sum_{i = 1}^{N_1} \ (x_i\rho_u) w_{ij}\Big).
           \label{eq:dropout_basis}
\end{align}
Looking at Equation \ref{eq:dropout_basis}, it can be seen that the
application of Dropout introduces stochasticity into the network
parameters in a similar manner as to that of the reparameterisation
trick shown in Equation~\ref{eq:reparam_example}. A key difference is
that in the case of Dropout, stochasticity is introduced into the
input space, as appose to the parameter space required for Bayesian
inference.  Yarin Gal \cite{gal2016uncertainty} identified this
similarity, and demonstrated how noise introduced through the
application of Dropout can be transferred to the networks weights efficiently as,
 \begin{align}
   \mb{W}^1_\rho& = \text{diag} (\bs{\rho}) \mb{W}^1
   \\
   \label{eq:mcMat}
   \mb{\Phi}_\rho &= a \big( \mb{X}^T  \mb{W}^1_\rho \big).
 \end{align}
Where $\bs{\rho}$ is a vector sampled from the Bernoulli distribution, and the $\text{diag}(\cdot)$ operator creates a square diagonal matrix from a vector. In doing this it can be seen that a single dropout variable is shared amongst each row of the weight matrix, allowing some correlation within rows to be maintained. By viewing the stochastic component in terms of network weights, the formulation becomes suitable for approximate inference using the VI framework. In this work, the approximate posterior is of the form of a Bernoulli
distribution multiplied by the network weights.
\par
The reparameterisation
trick is then applied to allow for partial derivatives w.r.t. network
parameters to be found. The ELBO is then formed and backpropagation is
performed to maximise the lower bound. MC integration is used to
approximate the analytically intractable expected log-likelihood. The
KL divergence between the approximate posterior and the prior
distribution in the ELBO is then found by approximating the scaled Bernoulli
approximate posterior as a mixture of two Gaussians with very small variance.
\par
In parallel to this work, Kingma \etal~\cite{kingma2015variational} identified this same similarity between
Dropout and it's potential for use within a VI framework. As appose to
the typical Bernoulli distributed r.v. introduced in Dropout,
\cite{kingma2015variational} focuses attention to the case when the
introduced r.v. is Gaussian \cite{wang2013fast}. It is also shown how with selection of an appropriate prior that is independent of parameters,
current applications of NNs using dropout can be viewed as approximate inference.
\par
Kingma \etal~also aims to reduce the variance in the stochastic gradients
using a refined, local reparameterisation. This is done by instead of
sampling from the weight distribution before applying the affine
transformation, the sampling is performed afterwards. For example,
consider a MFVB case where each weight is assumed to be an independent
Gaussian $w_{ij} \sim \N (\mu_{ij}, \sigma_{ij}^2)$. After the affine
transformation $\phi_j = \sum_{i = 1}^{N_1} (x_i\rho_i) w_{ij}$, the
posterior distribution of $\phi_j$ conditional on the inputs will also
be a factorised Gaussian,
\begin{align}
  q(\phi_j|\mb{x}) = \N (\gamma_j, \delta_j^2),
  \\
  \gamma_j = \sum_{i = 1}^N x_i \mu_{i,j},
  \\
  \delta_j^2 = \sum_{i = 1}^N x_i^2 \sigma_{i,j}^2.
\end{align}
It is advantageous to sample from this distribution for $\phi$ as
appose to the distribution of the weights $w$ themselves, as this
results in a gradient estimator whose variance scales linearly with
the number of mini-batches used during training.\footnote{This method
  also has computational advantages, as the dimension of $\mb{\phi}$ is
  typically much lower than that of $\bs{\omega}$.}
\par
These few bodies of work are important in addressing the serious lack
of rigour seen in ML research. For example, the initial Dropout paper
\cite{srivastava2014dropout} lacks any significant theoretical
foundation. Instead, the method cites a theory for sexual reproduction
\cite{livnat2010sex} as motivation for the method, and relies heavily
on the empirical results given. These empirical results have been
further demonstrated throughout many high impact\footnote{At the time
  of writing, \cite{srivastava2014dropout} has over ten thousand
  citations.} research items which utilise this technique merely as a
regularisation method. The work in \cite{gal2016uncertainty} and
\cite{kingma2015variational} show that there is theoretical
justification for such an approach. In attempts to reduce the effect
of overfitting in a network, the frequentist methodology relied on the
application of a weakly justified technique that shows empirical
success, while Bayesian analysis provides a rich body of theory that
naturally leads to a meaningful understanding of this powerful approximation.
\par
Whilst addressing the problem of applying VI to complex BNNs with more
hidden layers, practical implementations have shown inadequate
performance which is attributed to large variance in the MC
approximations of the gradient
computations. Hernandez \etal
\cite{hernandez2015probabilistic} acknowledge this limitation and
propose a new method for practical inference of BNNs titled
Probabilistic Back Propagation (PBP). PBP deviates from the typical VI
approach, and instead employs an Assumed Density Filtering (ADF)
method \cite{opper1998bayesian}. In this format, the posterior is
updated in an iterative fashion through application of Bayes rule,
\begin{equation}
  \label{eq:1}
  p(\bs{\omega}_{t+1} | \mathcal{D}_{t+1}) =
  \dfrac{ p(\bs{\omega}_{t} | \mathcal{D}_t) p(\mathcal{D}_{t+1}| \bs{\omega}_t)}{p(\mathcal{D}_{t+1})}.
\end{equation}
As opposed to traditional network training where the predicted error
is the objective function, PBP uses a forward pass to compute the
log-marginal probability of a target and updates the posterior
distribution of network parameters. The moment matching method defined
in \cite{minka2001family} updates the posterior using a variant of
backpropagation, whilst maintaining equivalent mean and variance
between the approximate and variational distribution,
\begin{align}
  \mu_{t+1} &= \mu_t + \sigma_t \dfrac{\partial \log p(\mathcal{D}_{t+1})}{ \partial \mu}
  \\
  \sigma_{t+1} &=
                 \sigma_t + \sigma_t^2 \Big[ \big( \dfrac{\partial \log p(\mathcal{D}_{t+1})}{\partial \mu_t} \big)^2 -
                 2 \dfrac{\partial \log p(\mathcal{D}_{t+1})}{ \partial \sigma} \Big].
\end{align}

Experimental results on multiple small data-sets illustrate reasonable
performance in terms of predicted accuracy and uncertainty estimation
when compared with HMC methods for simple regression
problems \cite{hernandez2015probabilistic}. A key limitation of this
method is the computational bottleneck introduced by the online
training method. This approach may be suitable for some applications,
or for updating existing BNNs with additional data as it becomes
available, though for performing inference on large data sets the
method is computationally prohibitive.
\par
A promising method for approximate inference in BNNs was proposed by
Blundell \etal, titled ``Bayes by Backprop''~\cite{blundell2015weight}. The method utilises the reparameterisation
trick to show how unbiased estimates of the derivative of an
expectation can be found. For a random variable
$ \bs{\omega} \sim q_{\bs{\theta}}(\bs{\omega})$ that can be
reparameterised as deterministic and differentiable function
$ \bs{\omega} = g(\bs{\epsilon}, \bs{\theta})$, the derivative of the
expectation of an arbitrary function $f(\bs{\omega}, \bs{\theta})$ can
be expressed as,
\begin{align}
  \dfrac{\partial}{\partial \bs{\theta}} \mathbb{E}_{q} [ f(\bs{\omega}, \bs{\theta}) ]
  &=
    \dfrac{\partial}{\partial \bs{\theta}} \int q_{\bs{\theta}}(\bs{\omega}) f(\bs{\omega}, \bs{\theta}) d\bs{\omega}
  \\
  &= \dfrac{\partial}{\partial \bs{\theta}} \int p(\bs{\epsilon}) f(\bs{\omega}, \bs{\theta}) d\bs{\epsilon}
  \\
  \label{eq:bbbGrad}
  &= \mathbb{E}_{q(\epsilon)} \Big[ \dfrac{\partial f(\bs{\omega}, \bs{\theta})}{\partial \bs{\omega}} \dfrac{\partial \bs{\omega}}{\partial \bs{\theta}} + \dfrac{\partial f(\bs{\omega}, \bs{\theta})}{\partial \bs{\theta}} \Big].
\end{align}
\par
In the Bayes by Backprop algorithm, the function
$ f(\bs{\omega}, \bs{\theta}) $ is set as,
\begin{equation}
  \label{eq:bbb}
  f(\bs{\omega}, \bs{\theta}) = \log \dfrac{q_{\bs{\theta}}(\bs{\omega})}{p(\bs{\omega})} - \log p(\mb{X}|\bs{\omega}).
\end{equation}
This $f(\bs{\omega}, \bs{\theta})$ can be seen as the argument for the
expectation performed in Equation~\ref{eq:KLRe}, which is part of the lower
bound.
\par
Combining Equations~\ref{eq:bbbGrad} and\ref{eq:bbb},
\begin{equation}
  \label{eq:bbbCost}
  \mathcal{L}(\bs{\omega}, \bs{\theta}) = \mathbb{E}_{q} [ f(\bs{\omega}, \bs{\theta}) ] = \mathbb{E}_{q} \Bigg[ \log \dfrac{q_{\bs{\theta}}(\bs{\omega})}{p(\bs{\omega})} - \log p(\mathcal{D}|\bs{\omega}) \Bigg] = - \mathcal{F}[q_{\bs{\theta}}]
\end{equation}
which is shown to be the negative of the ELBO, meaning that Bayes by
Backprop aims to minimise the KL divergence between the approximate
and true posterior. Monte Carlo integration is used\footnote{Some
  terms may be tractable in this integrand, depending on the form of
  the prior and posterior approximation. MC integration allows for
  arbitrary distributions to be approximated.} to approximate the cost
in Equation~\ref{eq:bbbCost},
\begin{equation}
  \label{eq:bbbApprox}
  \mathcal{F}[q_{\bs{\theta}}] \approx \sum_{i=1}^{N}
  \log \dfrac{q_{\bs{\theta}}(\bs{\omega}_i)}{p(\bs{\omega}_i)} - \log p(\mb{X}|\bs{\omega}_i)
\end{equation}
where $\bs{\omega}_i$ is the $i^{th}$ sample from
$q_{\bs{\theta}}(\bs{\omega})$. With the approximation in Equation~\ref{eq:bbbApprox}, the unbiased gradients can be found using the
result shown in Equation~\ref{eq:bbbGrad}.
\par
For the Bayes by Backprop algorithm, a fully factorised Gaussian
posterior is assumed such that $\bs{\theta} = \{ \bs{\mu}, \bs{\rho} \}$,
where $\bs{\sigma} = \text{softplus}(\bs{\rho})$ is used to ensure the
standard deviation parameter is positive. With this, the distribution of weights
$\bs{\omega} \sim \N(\bs{\mu}, \text{softplus}(\bs{\rho})^2)$ in the
network are reparameterised as,
\begin{equation}
  \bs{\omega} = g(\bs{\theta}, \bs{\epsilon}) = \mu + \text{softplus}(\bs{\rho}) \odot \bs{\epsilon}.
\end{equation}
In this BNN, the trainable parameters are $\bs{\mu}$ and
$\bs{\rho}$. Since a fully factorised distribution is used, following
from Equation~\ref{eq:factor}, the logarithm of the approximate posterior can
be represented as,
\begin{equation}
  \label{eq:factorLog}
  \log q_{\bs{\theta}}(\bs{\omega}) = \sum_{l,j,k} \log\Big( \mathcal{N}( w_{ljk}; \mu_{ljk}, \sigma_{ljk}^2) \Big).
\end{equation}
The complete Bayes by Backprop algorithm is described in Algorithm~\ref{alg:bbb}.
\begin{algorithm}
  \caption{Bayes by Backprop (BbB) algorithm \cite{blundell2015weight}}
  \label{alg:bbb}
  \begin{algorithmic}[1]
    \Procedure{BbB}{$\bs{\theta}, \mb{X}, \alpha$} \Repeat
    \State{$\mathcal{F}[q_{\bs{\theta}}] \gets
      0$}\Comment{Initialise cost} \For{$i$ in $[1, ..., N]$}
    \Comment{Number of samples for MC estimate} \State{Sample
      $\bs{\epsilon}_i \sim \N(\mb{0},\mb{1})$}
    \State{$\bs{\omega} \gets \bs{\mu} + \text{softplus}(\bs{\rho})
      \cdot \bs{\epsilon}_i$}
    \State{$ \mathcal{L} \gets \log q(\bs{\omega}|\bs{\theta}) - \log
      p(\bs{\omega}) - \log p(\mb{X}|\bs{\omega})$}
    \State{$\mathcal{F}[q_{\bs{\theta}}] +=
      \text{sum}(\mathcal{L}) / N$} \Comment{Sum across all log of
      weights in set $\bs{\omega}$} \EndFor
    \State{$\bs{\theta} \gets \bs{\theta} - \alpha \nabla_\theta
      \mathcal{F}[q_{\bs{\theta}}]$} \Comment{Update
      parameters} \Until{convergence} \EndProcedure
  \end{algorithmic}
\end{algorithm}
\subsection{Gaussian Process Properties of BNNs}
\begin{figure}[!h]
  \centering
  \subfloat[][]{\includegraphics[width=0.5\linewidth]{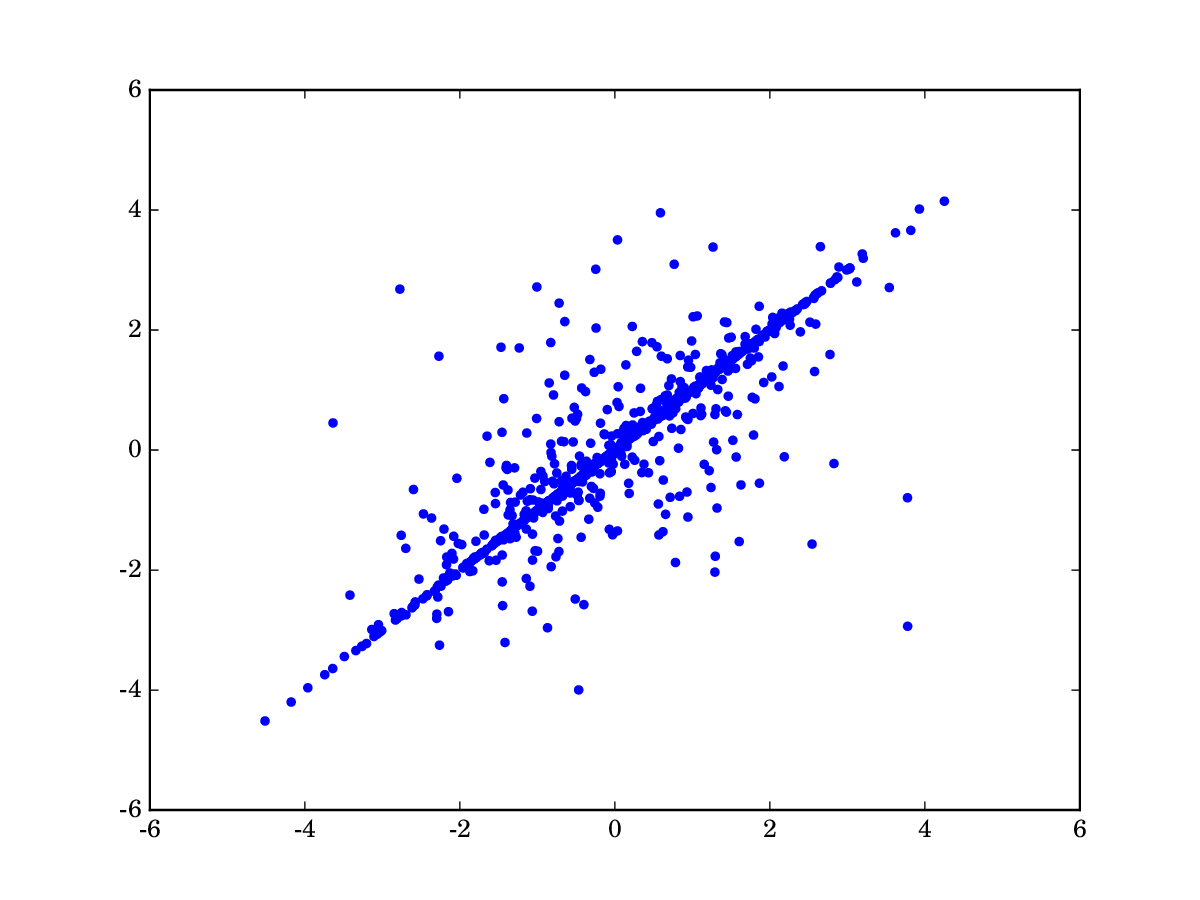}}
  \subfloat[][]{\includegraphics[width=0.5\linewidth]{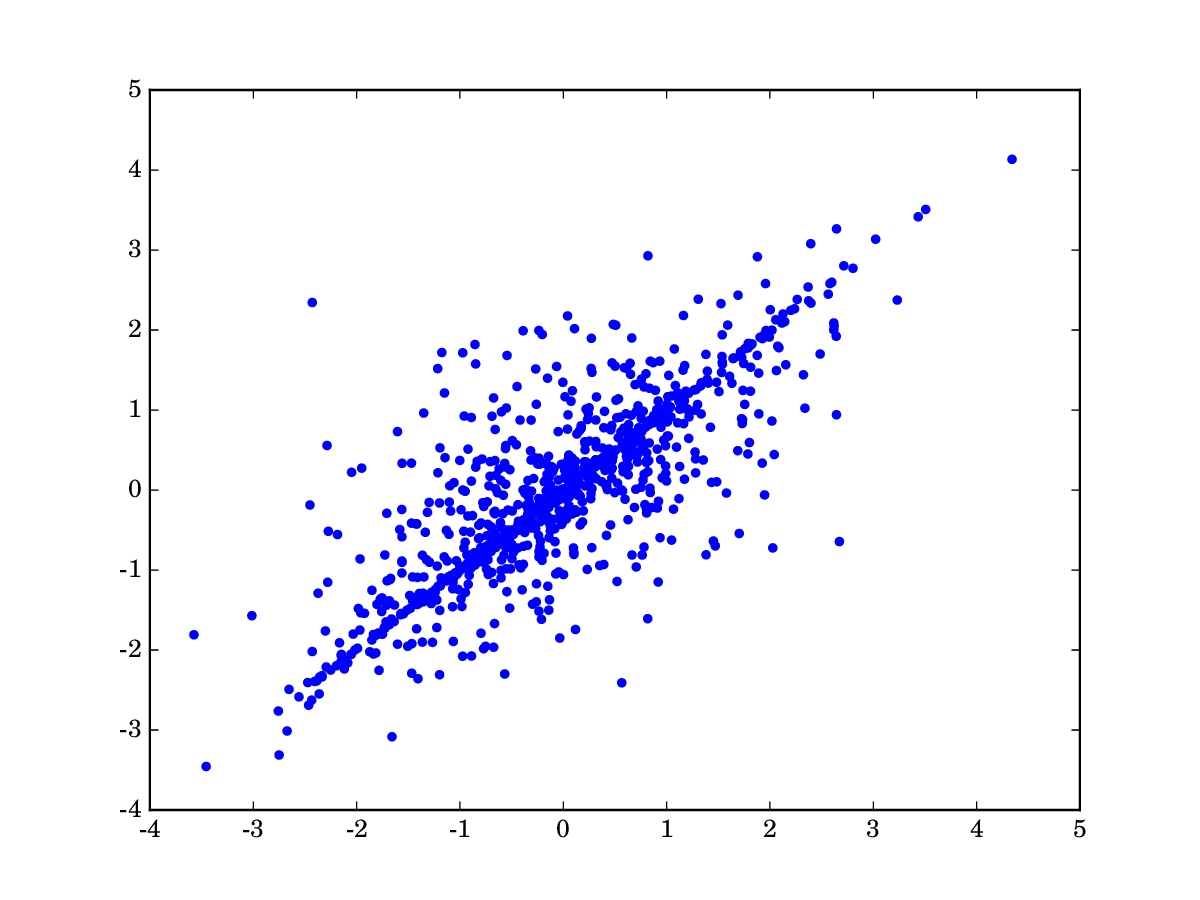}}
  \\
  \subfloat[][]{\includegraphics[width=0.5\linewidth]{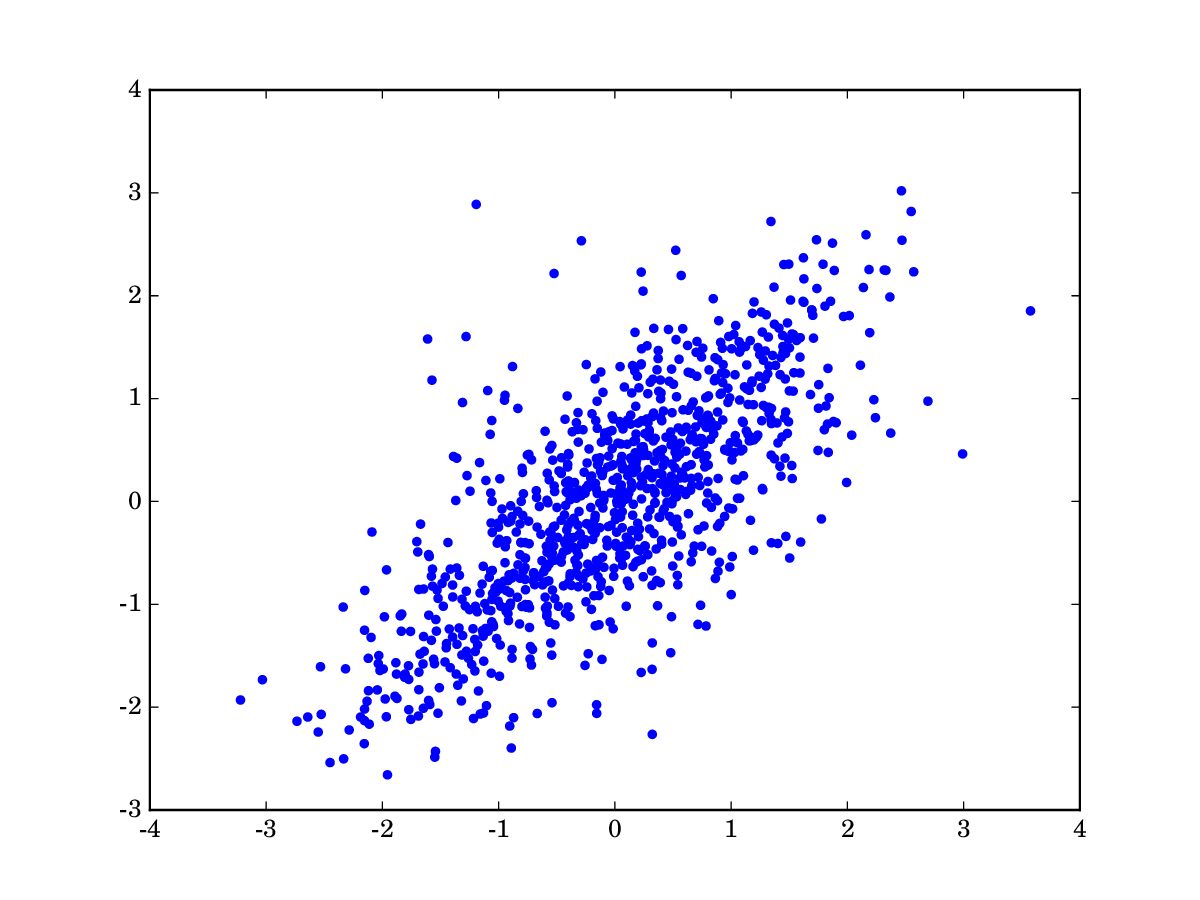}}
  \subfloat[][]{\includegraphics[width=0.5\linewidth]{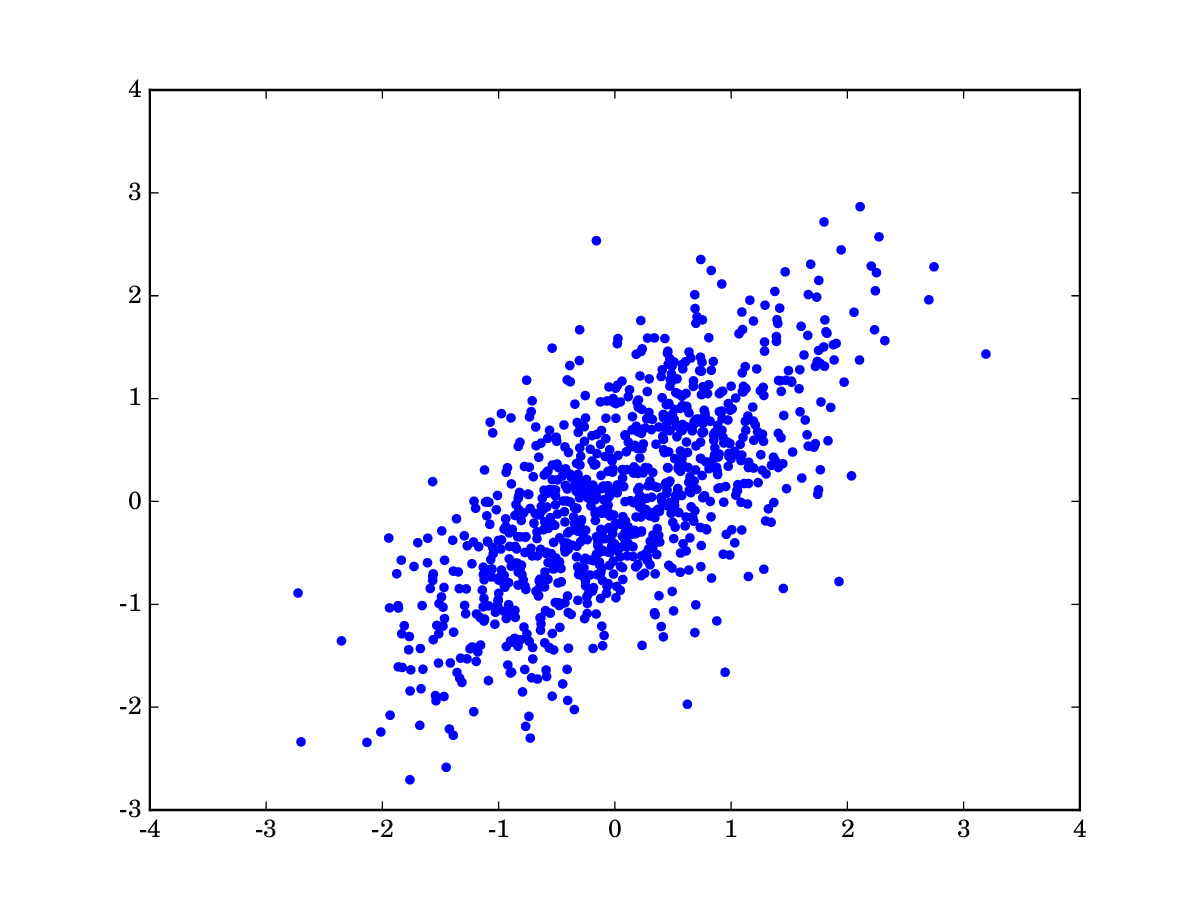}}
  \caption{Illustration of GP prior induced on output when placing a
    Gaussian prior over parameters as the network size
    increases. Experimentation replicated from
    \cite[p.~33]{neal1996bayesian}. Each dot corresponds to the output
    of a network with parameters sampled from the prior, with the
    x-axis as $f(0.2)$ and the y-axis as $f(-0.4)$. For each network,
    the number of hidden units are (a) 1, (b) 3, (c) 10, (d) 100.}
  \label{fig:prior}
\end{figure}

Neal~\cite{neal1996bayesian} also provided derivation and
experimentation results to illustrate that for a network with a single
hidden layer, a Gaussian Process (GP) prior over the network output
arises when the number of hidden units approaches infinity, and a
Gaussian prior is placed over parameters\footnote{For a regression
  model with no non-linear activation function placed on the output
  units.}. Figure \ref{fig:prior} illustrates this result.
\par
\par
This important link between NNs and GPs can be seen from Equations~\ref{eq:basis} and \ref{eq:mlp}. From these expressions, it can be seen
that a NN with a single hidden layer is a sum of $N$ parametric basis
functions applied to the input data. If the parameters for each basis
function in Equation~\ref{eq:basis} are r.v.'s, Equation~\ref{eq:mlp} becomes
the sum of r.v.'s. Under the central limit theorem, as the number of
hidden layers $N \rightarrow \infty$, the output becomes
Gaussian. Since the output is then described as an infinite sum of
basis functions, the output can be seen to become a GP. Following from
a full derivation of this result and the illustrations show in Figure
\ref{fig:prior},~\cite{neal1996bayesian} shows how an approximate
Gaussian nature is achieved for finite computing resources and how the
magnitude of this sum can be
maintained. Williams then demonstrated how
the form of the covariance function could be analysed for different
activation functions~\cite{williams1997computing}. The relation between GPs and infinitely wide networks with a single hidden layer work has recently been extended to the case
of deep networks~\cite{lee2018deep}.
\par
Identification of this link has motivated many research works in
BNNs. GPs provide many of the properties we wish to obtain, such as
reliable uncertainty estimates, interpretability and robustness. GPs
deliver these benefits at the cost of predictive performance and
exponentially large computational resources required as the size of
data sets increase. This link between GPs and BNNs has motivated the
merging of the two modelling schemes; maintaining the predictive
performance and flexibility seen in NNs while incorporating the
robustness and probabilistic properties enabled by GPs. This has led
to the development of the Deep Gaussian Process.
\par
Deep GPs are a cascade of individual GPs, where much like a NN, the
output of the previous GP serves as the input to a new GP
\cite{damianou2013deep, damianou2015deep}. This stacking of GPs allows
for learning of non-Gaussian densities from a combination of
GPs\footnote{A complete introduction to Deep GPs, along with code and
  lectures has been offered by Neil Lawrence
  \cite{lawrence2019deep}.}. A key challenge with GPs is fitting to
large data sets, as the dimensions of the Gram matrix for a single GP
is quadratic with the number of data points. This issue is amplified
with a Deep GP, as each individual GP in the cascade induces an
independent Gram matrix. Furthermore, the marginal likelihood for Deep
GPs are analytically intractable due to non-linearities in the
functions produced. Building on the work in \cite{damianou2011variational},
Damianou and Lawrence~\cite{damianou2013deep} use a VI approach to create an approximation that is tractable and
reduces computational complexity to that typically seen in sparse GPs
\cite{titsias2009variational}.
\par
Deep GPs have shown how the GPs can benefit from methodology seen in
NNs. Gal and Ghahramani \cite{gal2015dropoutslides, gal2016dropout,
  gal2015dropoutAppendix} built of this work to show how a Deep GP can
be approximated with a BNN\footnote{Approximation becomes a Deep GP as
  the number of hidden units in each layer approaches
  $\infty$.}. %
This is an expected result; given that Neal~\cite{neal1996bayesian} identified an infinitely wide network with a single hidden layer converges to a Gaussian process, by concatenating multiple infinitely wide layers we converge to a deep Gaussian process.
\par
Alongside this analysis of deep Gaussian processes, \cite{gal2015dropoutslides, gal2016dropout,
  gal2015dropoutAppendix} build on the work in \cite{williams1997computing} to analyse the relationship between the modern non-linear activation used within
BNNs and the covariance function for a GP. This is promising work that
could allow for more principled selection of activation functions in
NNs, similar to that of GPs. Which activation functions will yield a
stationary process? What is the expected length scale for our process?
These questions may be able to be addressed using the rich theory
existing for GPs.
\par
The GP properties are not restricted to MLP BNNs. Recent research has
identified certain relationships and conditions that induce GP
properties in convolutional BNNs \cite{garriga2018deep, novak2019bayesian}. This result
is expected since CNNs can be implemented as MLPs with structure
enforced in the weights. What this work identifies is how the GP is
constructed when this structure is
enforced. Van der Wilk \etal
\cite{van2017convolutional} proposed the Convolutional Gaussian
Process, which implements a patch based operation similar to that seen
in CNNs to define the GP prior over functions. Practical
implementation of this method requires the use of approximation
methods, due to the prohibitive cost of evaluating large data sets,
and even evaluation at each patch. Inducing points are formed with a
VI framework to reduce the number of data points to evaluate and the
number of patches evaluated.
\subsection{Limitations in Current BNNs}
Whilst great effort has been put into developing Bayesian methods for performing inference in NNs, there are significant limitations to these methods and many gaps remaining in the literature. A key limitation is the heavy reliance on VI methods. Within the VI framework, the most common approach is the Mean Field approach. MFVB provides a convenient way to represent an approximate posterior distribution by enforcing strong assumptions of independence between parameters. This assumption allows for factorised distributions to be used to approximate the posterior. This assumption of independence significantly reduces the computational complexity of approximate inference at the cost of probabilistic accuracy.
\par
A common finding with VI approaches is that resulting models are overconfident, in that predictive means can be accurate while variance is considerably under estimated \cite{mackay2003information, wang2005inadequacy, turner2011two, blei2017variational, giordana2018covariances}. This phenomenon is described in Section 10.1.2 of \cite{bishop2006} and Section 21.2.2 of \cite{murphey2012machine}, both of which are accompanied by examples and intuitive figures to illustrate this property. This property of under-estimated variance is present within much of the current research in BNNs \cite{gal2016uncertainty}. Recent work has aimed to address these issues through the use of noise contrastive priors \cite{hafner2018reliable} and through use of calibration data sets \cite{kuleshov2018accurate}. The authors in \cite{gal2017concrete} employ the use of the concrete distribution \cite{maddison2016concrete} to approximate the Bernoulli parameter in the MC Dropout method \cite{gal2016dropout}, allowing for it to be optimised, resulting in posterior variances that are better calibrated. Despite these efforts, the task of formulating reliable and calibrated uncertainty estimates within a VI framework for BNNs remains unsolved.
\par
It is reasonable to consider that perhaps the limitations of the current VI approaches are influenced by the choice of approximate distribution used, particularly the usual MFVB approach of independent Gaussians. If more comprehensive approximate distributions are used, will our predictions be more consistent with the data we have and haven't seen? Mixture based approximations have been proposed for the general VI approach \cite{jaakkola1998improving, jordan1999introduction}, though introduction of $N$ mixtures increases the number of variational parameters by $N$. Matrix-Normal approximate posteriors have been introduced to the case of BNNs \cite{louizos2016structured}, which reduces the number of variational parameters in the model when compared with a full rank Gaussian, though this work still factorises over individual weights, meaning no covariance structure is modelled\footnote{Though this work highlights that even with a fully factorised distribution over weights, the outputs of each layer will be correlated.}. MCDropout is able to maintain some correlation information within the rows of weight matrix, at the compromise of a low entropy approximate posterior.
\par
A recent approach for VI has been proposed to capture more complex posterior distributions through the use of normalising flows \cite{tabak2010, tabak2013family}. Within a normalising flow, the initial distribution ``flows'' through a sequence of invertible functions to produce a more complex distribution. This can be applied within the VI framework using amortized inference \cite{rezende2015variational}. Amortized inference introduces an inference network which maps input data to the variational parameters of generative model. These parameters are then used to sample from the posterior of the generative process. The use of normalising flows has been extended to the case of BNNs \cite{louizos2017multiplicative}. Issues arise with this approach relating to the computational complexity, along with limitations of amortized inference. Normalising flows requires the calculation of the determinant of the Jacobian for applying the change of variables used for each invertible function, which can be computationally expensive for certain models. Computational complexity can be reduced by restricting the normalising flow to contain invertible operations that are numerically stable \cite{rezende2015variational, dinh2016density}. These restrictions have been shown to severely limit the flexibility of the inference process, and the complexity of the resulting posterior approximation \cite{cremer2018inference}.
\par
As stated previously, in the VI framework, an approximate distribution is selected and the ELBO is then maximised. This ELBO arises from the applying the KL divergence between the true and approximate posterior, but this begs the question, why use the KL? The KL divergence is a well known measure to assess the similarity of between two distributions, and satisfies all the key properties of a divergence (ie. is positive and only zero when the two distributions are equal). A divergence allows us to know whether our approximation is approaching the true distribution, but not how close we are to it. Why not use of a well defined distance as appose to a divergence?
\par
The KL divergence is used as it allows us to separate the intractable quantity (the marginal likelihood) out of our objective function (the ELBO) which we can optimise. Our goal with our Bayesian inference is to identify the parameters that best fit our model under prior knowledge and the distribution of the observed data. The VI framework poses inference as an optimisation problem, where we optimise our parameters to minimise the KL divergence between our approximate and true distribution (which maximises our ELBO). Since we are optimising our parameters, by separating the marginal likelihood from our objective function, we are able to compute derivatives with respect to the tractable quantities. Since the marginal likelihood is independent of the parameters, this component vanishes when the derivative is taken. This is the key reason why the KL divergence is used, as it allows us to separate the intractable quantity out of our objective function, which will then be evaluated as zero when using gradient information to perform optimisation.
\par
The KL divergence has been shown to be part of a generic family of divergences known as $\alpha$-divergences \cite{amari2012differential, minka2005divergence}. The $\alpha$-divergence is represented as,
\begin{equation}
  \label{eq:alpha_d}
  D_\alpha[p(\omega) || q(\omega)] =
  \dfrac{1}{\alpha(1 - \alpha)} \Big( 1 - \int p(\omega)^\alpha q(\omega)^{1 - \alpha}d\omega \Big).
\end{equation}
The forward KL divergence used in VI is found from Equation~\ref{eq:alpha_d} in the limit that $\alpha \rightarrow 0$, and the reverse KL divergence $\text{KL}(p || q)$ occurs in the limit of $\alpha \rightarrow 1$, which is used during expectation propagation. While the use of the forward KL divergence used in VI typically results in an under-estimated variance, the use of the reverse KL will often over-estimate variance~\cite{bishop2006}. Similarly, the Hellinger distance arises from Eq. \ref{eq:alpha_d} when $\alpha = 1/2$,
\begin{equation}
  \label{eq:hellinger}
  D_H(p(\omega) || q(\omega))^2 = \frac{1}{2}\int \Big( p(\omega)^{\frac{1}{2}} - q(\omega)^{\frac{1}{2}} \Big)^2 d\omega.
\end{equation}
This (squared Hellinger distance) is a valid distance, in that it satisfies the triangle inequality and is symmetric. Minimisation of the Hellinger distance has shown to provide reasonable compromise in variance estimate when compared with the two KL divergences~\cite{minka2005divergence}. Though these measures may provide desirable qualities, they are not suitable for direct use within VI, as the intractable marginal likelihood cannot be separated from the other terms of interest\footnote{This may be easier to see for the Hellinger distance, but perhaps less so for the reverse KL divergence. Enthusiastic readers are encouraged to not take my word for it, and to put pen and paper to prove this for themselves!}. While these measures cannot be immediately used, it illustrates how a change in the objective measure can result in different approximations. It is possible that more accurate posterior expectations can be found by utilising a different measure for the objective function.
\par
The vast majority of modern works have revolved around the notion of VI. This is largely due to its amenability to SGD. Sophisticated tools now exist to simplify and accelerate the implementation of automatic differentiation and backpropagation~\cite{jia2014caffe, chollet2015, tensorflow2016, dillon2017tensorflow, paszke2017automatic, mxnet2015, kucukelbir2016automatic}. Another benefit of VI is it's acceptance of sub-sampling in the likelihood. Sub-sampling reduces the computational expense for performing inference required to train over large data sets currently available. It is this key reason that more traditional MCMC based methods have received significantly less attention in the BNN community.
\par
MCMC serves as the gold standard for performing Bayesian inference due to it's rich theoretical development, asymptotic guarantees and practical convergence diagnostics. Traditional MCMC based methods require sampling from the full joint likelihood to perform updates, requiring all training data to be seen before any new proposal can be made. Sub-sampling MCMC, or Stochastic Gradient MCMC (SG-MCMC) approaches have been proposed in \cite{welling2011sgld, patterson2013stochastic, chen2014hmc}, which have since been applied to BNNs \cite{chen2015preconditioned}. It has since been shown that the naive sub-sampling within MCMC will bias the trajectory of the stochastic updates away from the posterior \cite{betancourt2015fundamental}. This bias removes the theoretical advantages gained from a traditional MCMC approach, making them less desirable than a VI approach which is often less computationally expensive. For sampling methods to become feasible, sub-sampling methods need to be developed that assure convergence to the posterior distribution.

\section{Comparison of Modern BNNs}
From the literature survey presented within, two prominent methods for approximate inference in BNNs was Bayes by Backprop~\cite{blundell2015weight} and MC Dropout~\cite{gal2016dropout}. These methods have found to be the most
promising and highest impact methods for approximate inference in
BNNs. These are both VI methods that are flexible enough to permit the
use of SGD, making deployment to large and practical data sets
feasible. Given their prominence, it is worthwhile to compare the methods to see how well they perorm.
\par
To compare these methods, a series of simple homoskedastic regression tasks were conducted. For these regression models, the likelihood is represented as Gaussian. With this we can write that the un-normalised posterior is,
\begin{equation}
  \label{eq:nn_post}
  p(\bs{\omega} | \mathcal{D}) \propto p(\bs{\omega}) \mathcal{N}(\mb{f}^{\bs{\omega}}(\mathcal{D}), \sigma^2 \mb{I}),
\end{equation}
where $\mb{f}^{\bs{\omega}}(\mathcal{D})$ is the function represented by the BNNs. A mixture of Gaussians was used to model a spike-slab prior for both models. The approximate posterior $q_{\bs{\theta}}(\bs{\omega})$ was then found for each model using the respective methods proposed. For Bayes by Backprop, the approximate posterior is a fully factorised Gaussian, and for MC Dropout is a scaled Bernoulli distribution. With the approximate posterior for each model, predictive quantities can be found using MC Integration. The first two moments can be approximated as \cite{gal2016uncertainty},
\begin{align}
  \label{eq:firstmoment}
  \mathbb{E}_{q}[\mb{y}^*] &\approx \dfrac{1}{N} \sum_{i=1}^N \mb{f}^{\bs{\omega}_i}(\mb{x}^*)
  \\
  \label{eq:secondmoment}
  \mathbb{E}_{q}[\mb{y}^{*T}\mb{y}^{*}] &\approx \sigma^2 \mb{I} +
                                  \dfrac{1}{N} \sum_{i=1}^N \mb{f}^{\bs{\omega}_i}(\mb{x}^*)^T \mb{f}^{\bs{\omega}_i}(\mb{x}^*)
\end{align}
where the star superscript denotes the new input and output sample $\mb{x}^*, \mb{y}^*$ from the test set.
\par
\begin{figure}[!h]
  \subfloat{\includegraphics[width=0.33\linewidth]{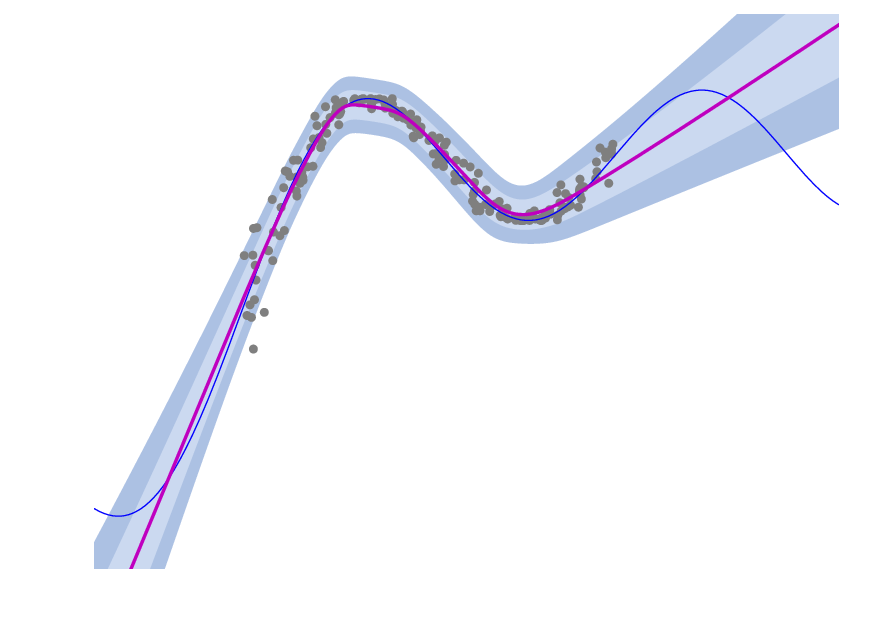}}
  \subfloat{\includegraphics[width=0.33\linewidth]{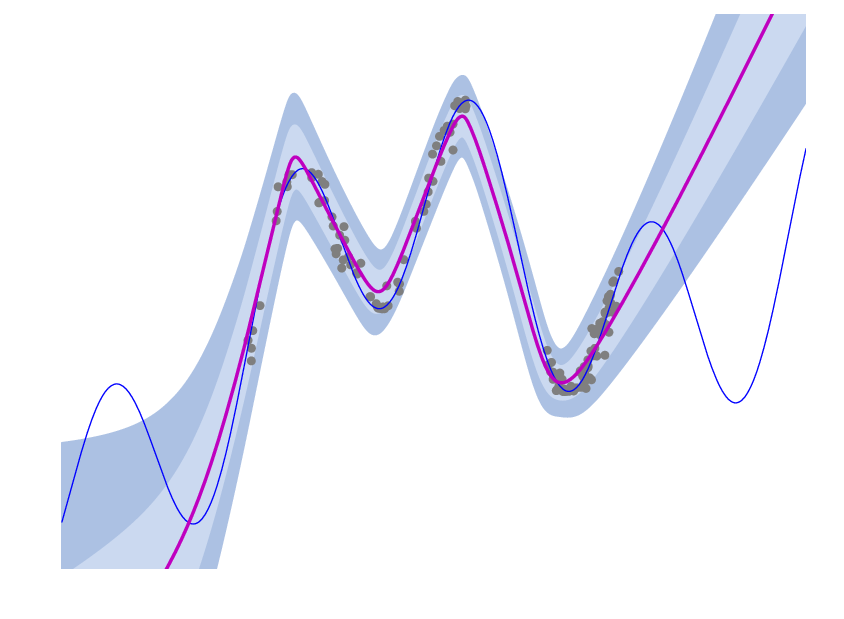}}
  \subfloat{\includegraphics[width=0.33\linewidth]{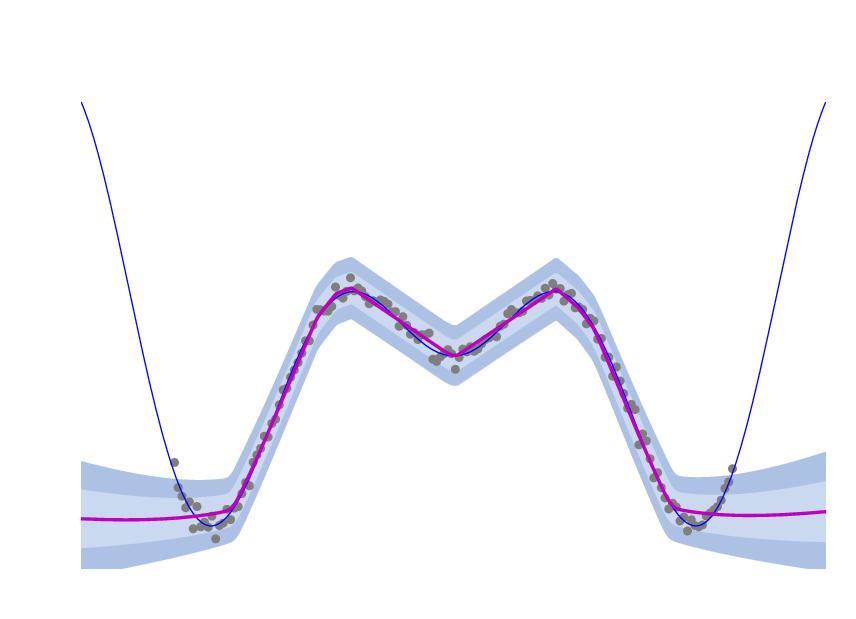}}
  \\
  \subfloat{\includegraphics[width=0.33\linewidth]{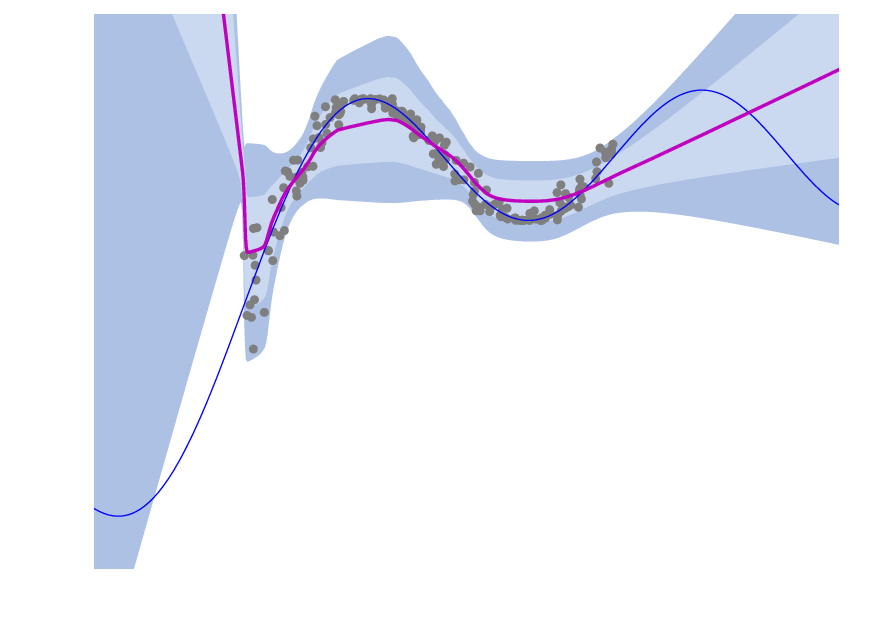}}
  \subfloat{\includegraphics[width=0.33\linewidth]{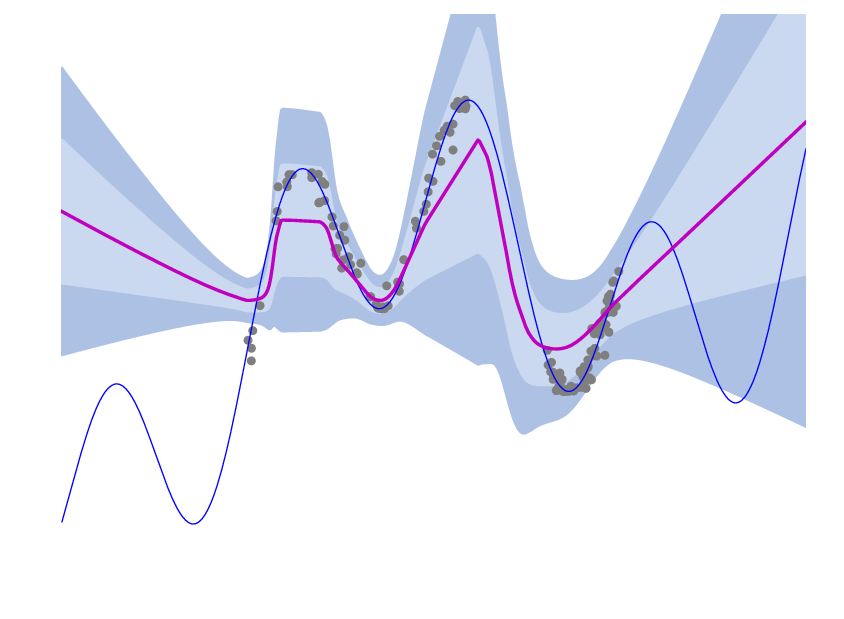}}
  \subfloat{\includegraphics[width=0.33\linewidth]{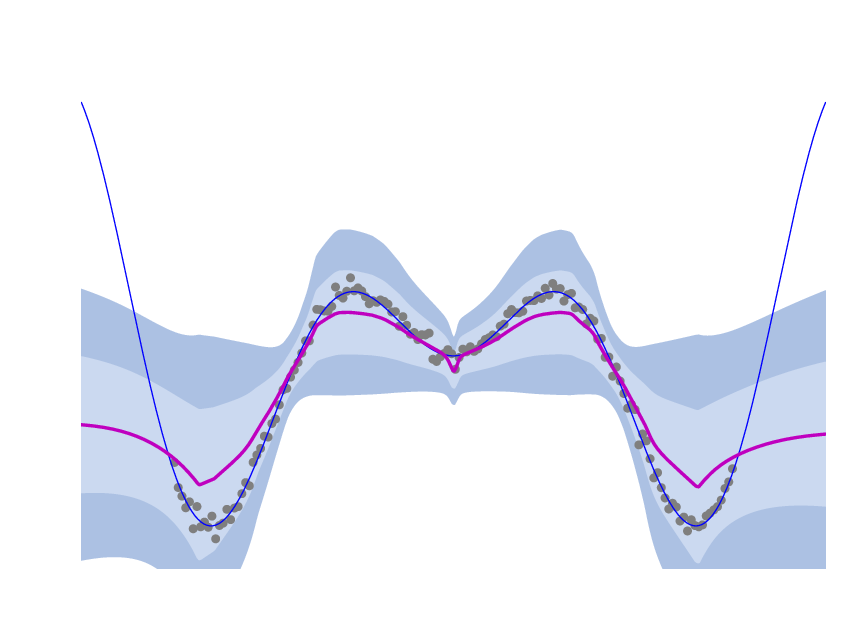}}
  \\
  \subfloat{\includegraphics[width=0.33\linewidth]{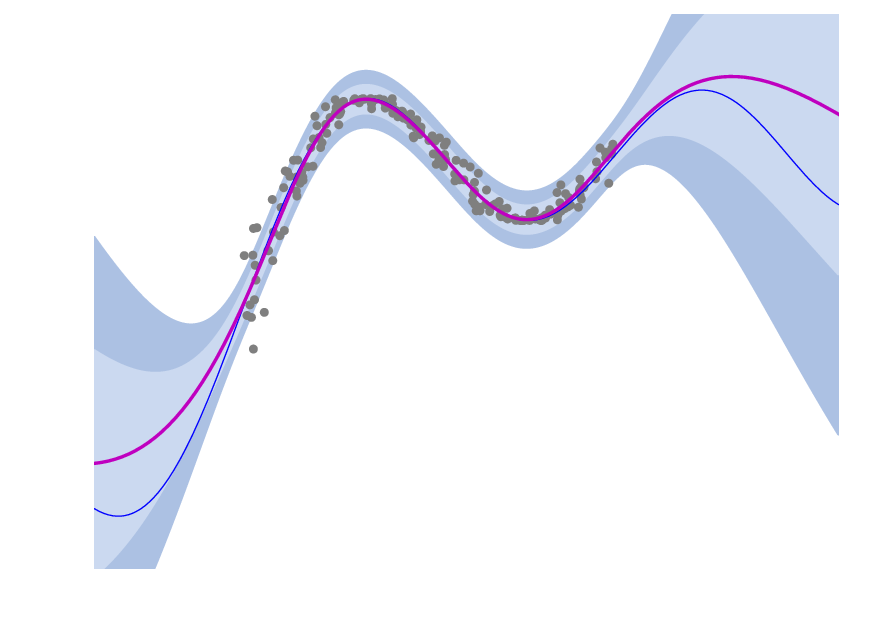}}
  \subfloat{\includegraphics[width=0.33\linewidth]{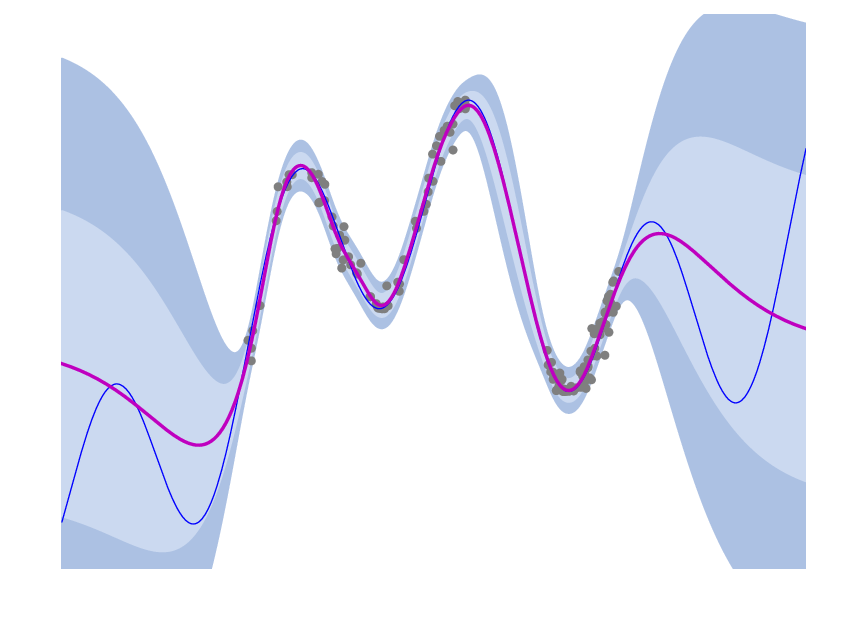}}
  \subfloat{\includegraphics[width=0.33\linewidth]{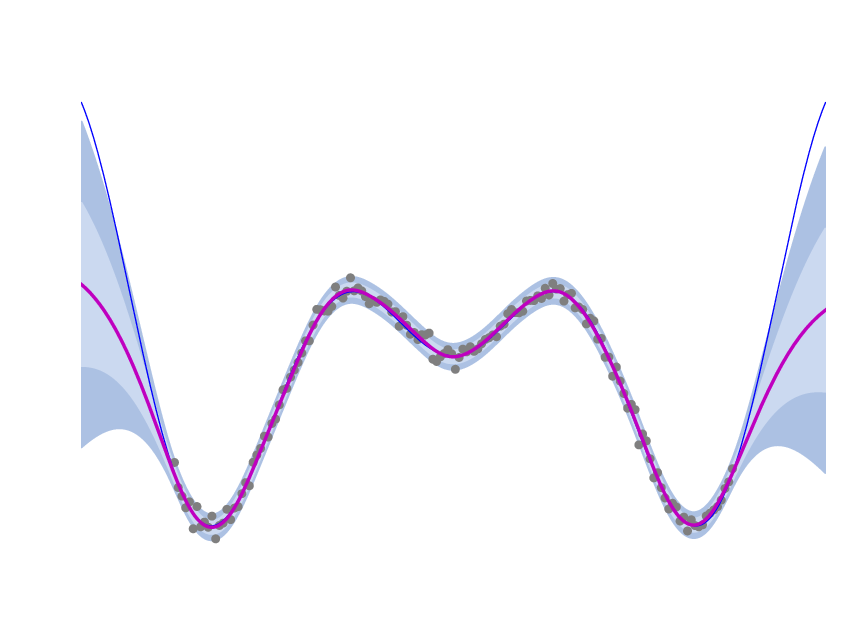}}
  \caption{Comparison of BNNs with GP for a regression task over
    three toy data sets. The top row is a BNN trained with Bayes By
    Backprop~\cite{blundell2015weight}, the centre row is trained
    with MC dropout \cite{gal2016uncertainty}, and the bottom a GP
    with a Mattern52 kernel fitted with the GPflow package~\cite{GPflow2017}. The two BNNs consisted of two hidden layers utilising ReLU activation. Training data is shown with the dark grey
    scatter, the mean is shown in purple, the true test function is
    shown in blue, and the shaded regions representing $\pm$ one and
    two std. from the mean. Best viewed on a computer screen.}
  \label{fig:reg}
\end{figure}

The data sets used to evaluate these models were simple toy data sets from high impact papers, where similar experimentation was provided as empirical evidence~\cite{blundell2015weight, osband2016deep}. Both BNN methods were then compared with a GP model.
Figure \ref{fig:reg} illustrates these results.

\par
Analysis of the regression results shown in Figure~\ref{fig:reg} shows
contrasting performance in terms of bias and variance in
predictions. Models trained with Bayes by Backprop and a factorised Gaussian
approximate posterior show reasonable predictive results withing the
distribution of training data, though variance outside the region of
training data is significantly under estimated when compared with the
GP. MC Dropout with a scaled Bernoulli approximate posterior typically
exhibits greater variance for out of distribution data, though maintains unnecessarily
high variance within the distribution of training data. Little tuning of hyperparameters was done to these models. Better results may be achieved, particularly for MC Dropout, with better selection of hyperparameters. Alternatively, a more complete Bayesian approach can be used, where hyperparameters are treated as latent variables and marginalisation is performed over these variables.
\par
It is worthwile noting the computational and practical difficulties encountered with these methods.
The MC Dropout method is incredibly versitle, in that it was less sensitive to the choice of prior distribution. It also managed to fit to more complex distributions with fewer samples and training iterations. On top all this is the significant savings in computational resources. Given that training a model using MC Dropout is often identical to how many existing deep networks are trained, inference is performed in the same time as traditional vanilla networks. It also offers no increase in the number of parameters to a network, where Bayes by Backprop requires twice as many. These factors should be taken into account for practical scenarios. If the data being modelled is smooth, is in sufficient quantity and additional time for inference is permitted, Bayes by Backprop may be preferable. For large networks with complex functions, sparse data and more stringent time requirements, MC Dropout may be more suitable.
\subsection{Convolutional BNNs}
\label{sec:convolutional-bnns}
Whilst the MLP serves as the basis for NNs, the most prominent NN architecture is the Convolutional Neural Network (CNN)~\cite{lecun1989backpropagation}. These networks have excelled at challenging image classification tasks, with predictive performance far exceeding prior kernel based or feature engineered methods. A CNN differs from a typical MLP through it's application a convolution-like operator as oppose to inner products\footnote{Emphasis is placed on ``convolution like'', as it is not equivalent to the mathematical operation of linear or circular convolution.}. The output of a single convolutional layer can be expressed as,
\begin{equation}
  \label{eq:4}
  \bs{\Phi} = u(\mathbf{X}^T \ast \mb{W})
\end{equation}
where $u(\cdot)$ is a non-linear activation and $\ast$ represents the convolution-like operation. Here the input $\mb{X}$ and the weight matrix $\mb{W}$ are no longer restricted to either vectors or matrices, and can instead be multi-dimensional arrays. It can be shown that CNNs can be written to have an equivalent MLP model, allowing for optimised linear algebra packages to be used for training with back-propagation \cite{goodfellow2016deep}.
\par
Extending on the current research methods, a new type of Bayesian Convolutional Neural Network (BCNN) can be developed. This is achieved here by extending on the Bayes by Backprop method~\cite{blundell2015weight} to the case of models suitable for image classification. Each weight in the convolutional layers is assumed to be independent, allowing for factorisation over each individual parameter.
\par
Experimentation was conducted to investigate the predictive performance of BCNNs, and the quality of their uncertainty estimates. These networks were configured for classification of the MNIST
hand digit dataset~\cite{lecun1998mnist}.
\par
Since this task is a classification task, the likelihood for the BCNN was set to a
Softmax function,
\begin{equation}
  \label{eq:softmax}
  \text{softmax}(\mb{f}_i^{\bs{\omega}}) = \dfrac{\exp \big( \mb{f}_i^{\bs{\omega}}(\mathcal{D}) \big)}{\sum_{j} \exp \Big(\mb{f}_j^{\bs{\omega}}(\mathcal{D})\Big)}.
\end{equation}
The un-normalised posterior can then be represented as,
\begin{equation}
  \label{eq:cnn_post}
  p(\bs{\omega} | \mathcal{D}) \propto p(\bs{\omega}) \times  \text{softmax}(\mb{f}^{\bs{\omega}}(\mathcal{D})).
\end{equation}
The approximate posterior is then found using Bayes by Backprop. Predictive mean for test samples can be found using Equation~\ref{eq:firstmoment}, and MC integration is used to approximate credible intervals~\cite{murphey2012machine}.
\par
Comparison with a vanilla CNN was made to evaluate the predictive performance of the BCNN. For both the vanilla and BCNN, the popular LeNet architecture~\cite{lecun1998mnist} was used. Classification was conducted using the mean output of the BCNN, with credible intervals being used to assess the models uncertainty. Overall predictive performance for both networks on the 10,000 test images in the MNIST dataset showed comparative performance. The BCNN showed a test prediction accuracy of 98.99\%, while the vanilla network showed a slight improvement with a prediction accuracy of 99.92\%. Whilst the competitive predictive performance is essential, the main benefit of the BCNN is that we yield valuable information about the uncertainty of our predictions. Examples of difficult to classify digits are shown in the Appendix, accompanied by plots of the mean prediction and 95\% credible intervals for each class. From these examples, we can see the large amount of predictive uncertainty for these challenging images, which could be used to make more informed decisions in practical scenarios.
\par

This uncertainty information is invaluable for many scenarios of interest. As statistical models are increasingly employed for complex tasks containing human interaction, it is crucial that many of these systems make responsible decisions based on their perceived model of the world. For example, NNs are largely used within the development of autonomous vehicles. Development of autonomous vehicles is an incredibly challenging feat, due to the high degree of variability in scenarios and the complexity relating to human interaction. Current technologies are insufficient for safely enabling this task, and as discussed earlier, the use of these technologies have been involved in multiple deaths \cite{tesla2016, abc2018}. It is not possible to model all variables within such a highly complex system. This accompanied by imperfect models and reliance on approximate inference, it is important that our models can communicate any uncertainty relating to decisions made. It is crucial that we acknowledge that in essence, our models are wrong. This is why probabilistic models are favoured for such scenarios; there is an underlying theory to help us deal with heterogeneity in our data and to account for uncertainty induced by variables not included in the model. It is vital that models used for such complex scenarios can communicate their uncertainty when used in such complex and high risk scenarios.

\section{Conclusion}
Throughout this report, the problems that arise with overconfident predictions from typical NNs and ad hoc model design have been illustrated. Bayesian analysis has been shown to provide a rich body of theory to address these challenges, though exact computation remains analytically and computationally intractable for any BNN of interest. In practice, approximate inference must be relied upon to yield accurate approximations to the posterior.
\par
Many of the approximate methods for inference within BNNs have revolved around the MFVB approach. This provides a tractable lower bound to optimise w.r.t variational parameters. These methods are attractive due to their relative ease of use, accuracy of predictive mean values and acceptable number of induced parameters. Despite this, it was shown through the literature survey and experimentation results that the assumptions made within a fully factorised MFVB approach result in over-confident predictions. It was shown that these MFVB approaches can be extended upon to more complex models such as CNNs. Experimental results indicate comparable predictive performance to point estimate CNNs for image classification tasks. The Bayesian CNN was able to provide credible intervals on the predictions, which were found to be highly informative and intuitive measure of uncertainty for difficult to classify data points.
\par
This review and these experiments highlight the capabilities of Bayesian analysis to address common challenges seen in the machine learning community. These results also highlight how current approximate inference methods for BNNs are insufficient and can provide inaccurate variance information. Additional research is required to not only determine how these networks operate, but how accurate inference can be achieved with modern large networks. Methods to scale exact inference methods such as MCMC to large data sets would allow for a more principled method of performing inference. MCMC offers diagnostic methods to assess convergence and quality of inference. Similar diagnostics for VI would allow researchers and practitioners to evaluate the quality of their assumed posterior, and inform them with ways to improve on this assumption. Achieving these goals will allow us to obtain accurate posterior approximations. From this we will be able to sufficiently determine what our models know, but also what they don't know.

\bibliographystyle{IEEEtran}%
\bibliography{references.bib}
\newpage
\section*{Appendix}
\label{ap:mnist}
\begin{figure*}[!h]
  \centering
  \subfloat[][]{\includegraphics[width=0.33\linewidth]{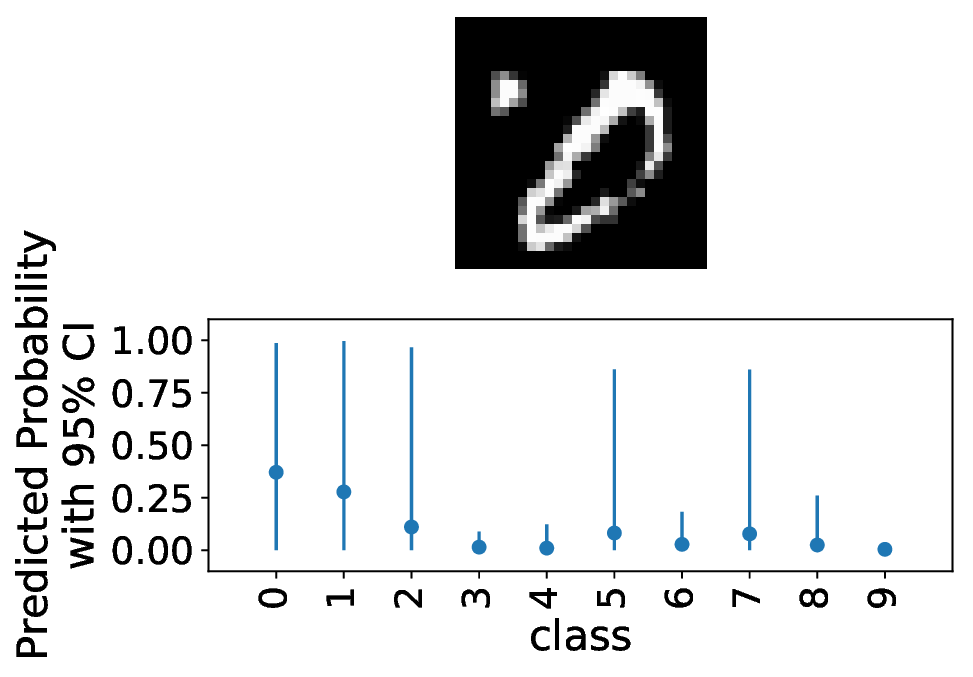}}
  \subfloat[][]{\includegraphics[width=0.33\linewidth]{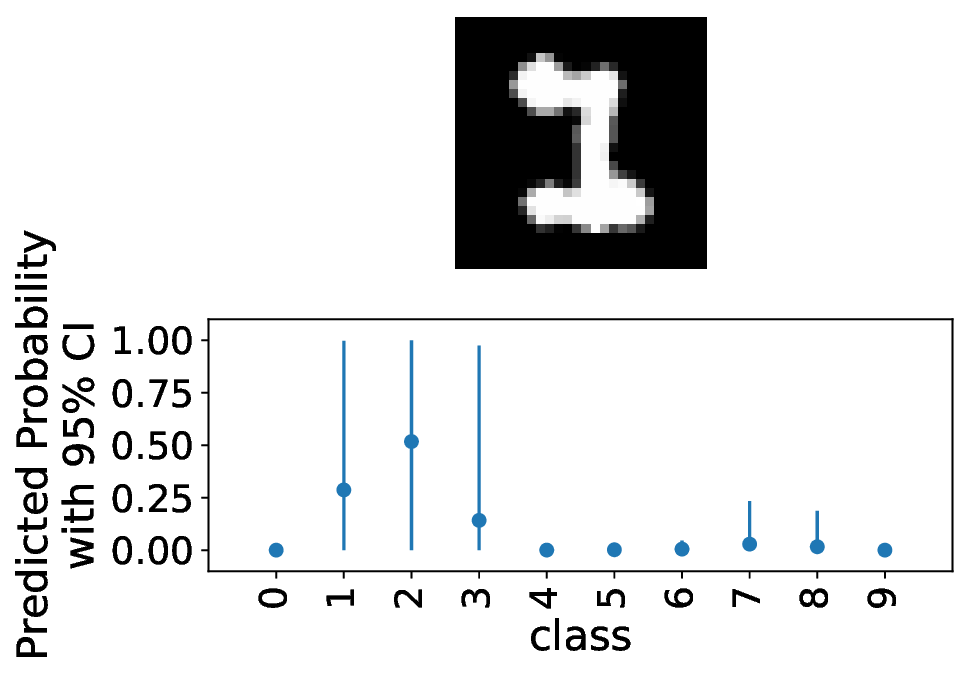}}
  \subfloat[][]{\includegraphics[width=0.33\linewidth]{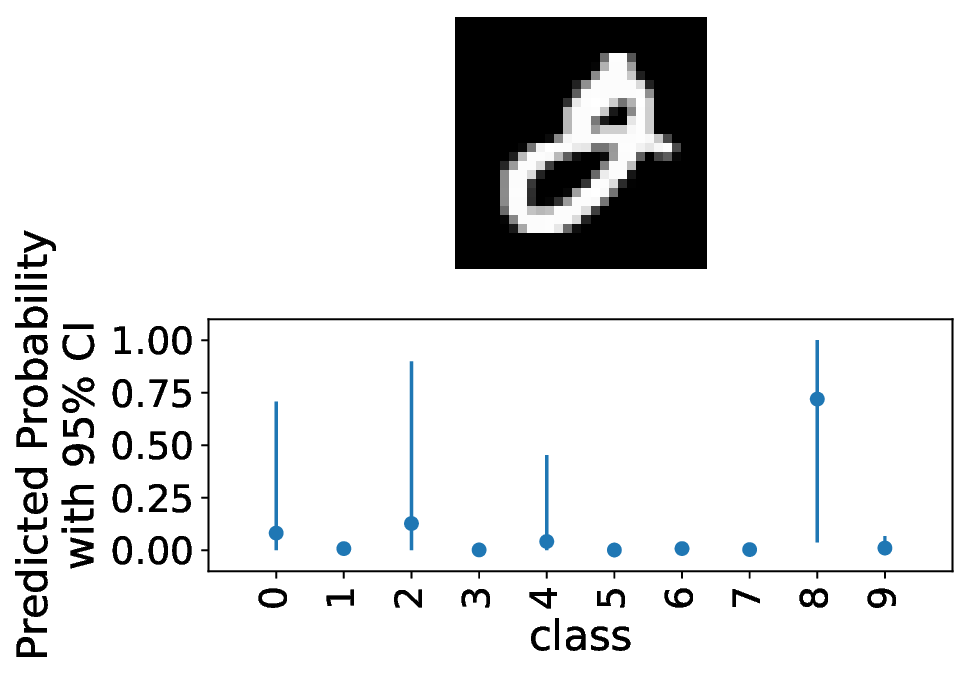}}
  \\
  \subfloat[][]{\includegraphics[width=0.33\linewidth]{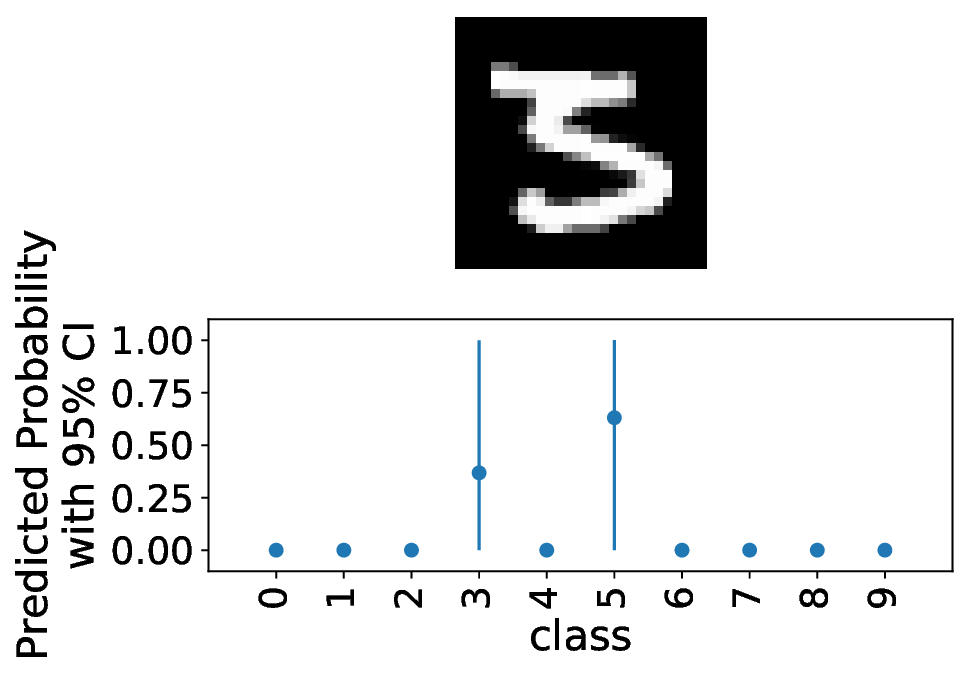}}
  \subfloat[][]{\includegraphics[width=0.33\linewidth]{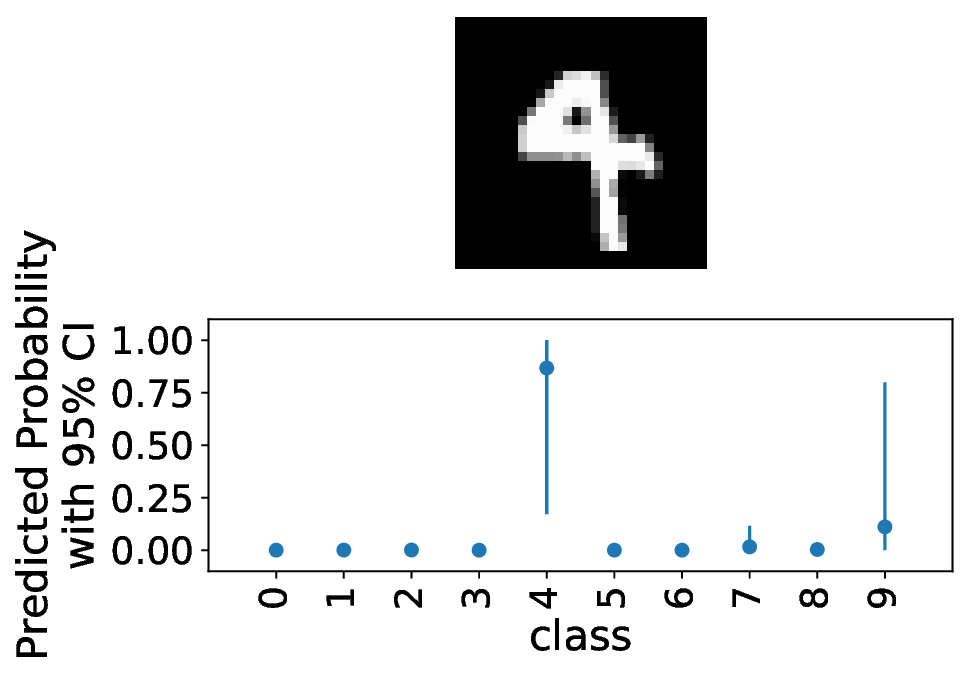}}
  \subfloat[][]{\includegraphics[width=0.33\linewidth]{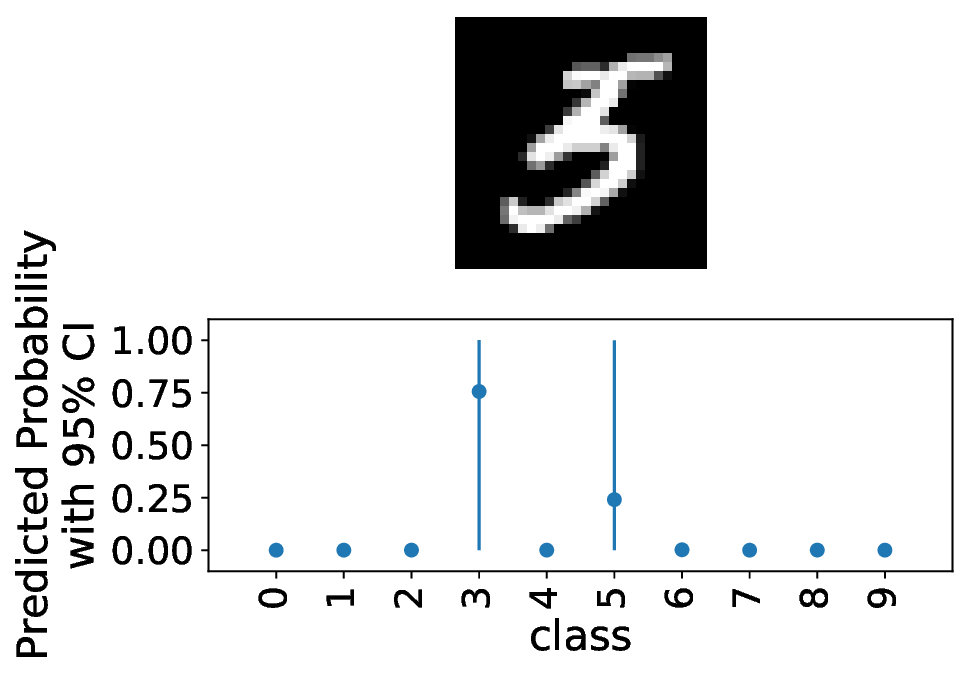}}
  \\
  \subfloat[][]{\includegraphics[width=0.33\linewidth]{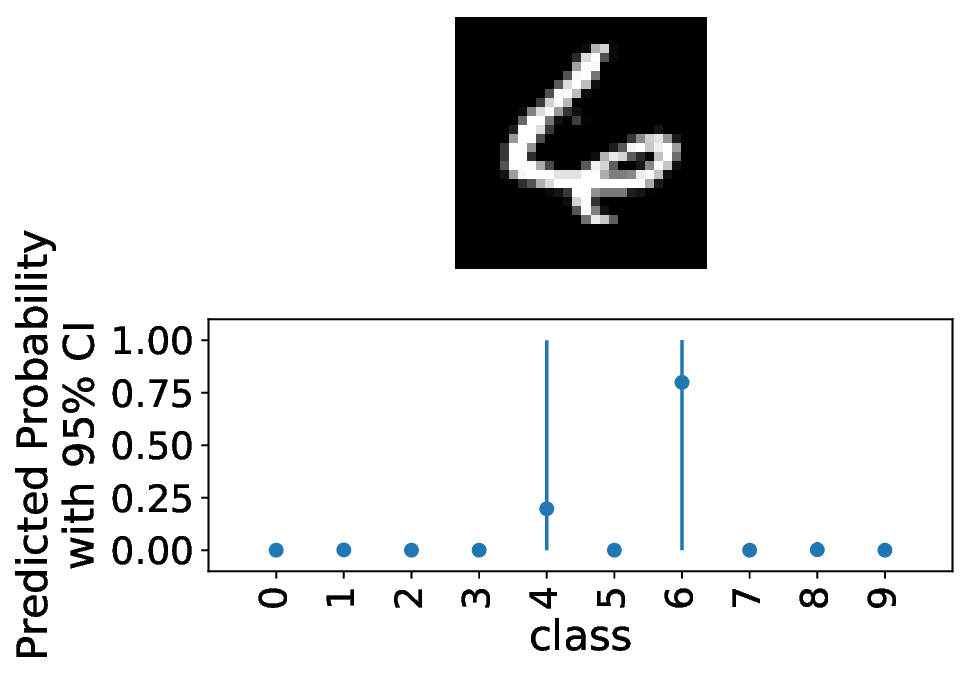}}
  \subfloat[][]{\includegraphics[width=0.33\linewidth]{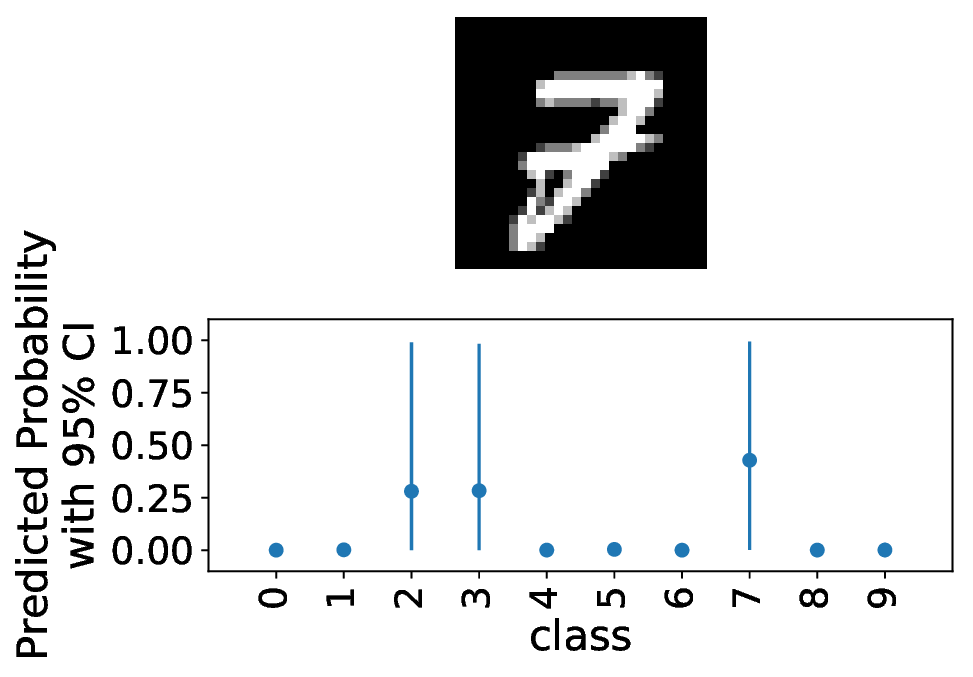}}
  \subfloat[][]{\includegraphics[width=0.33\linewidth]{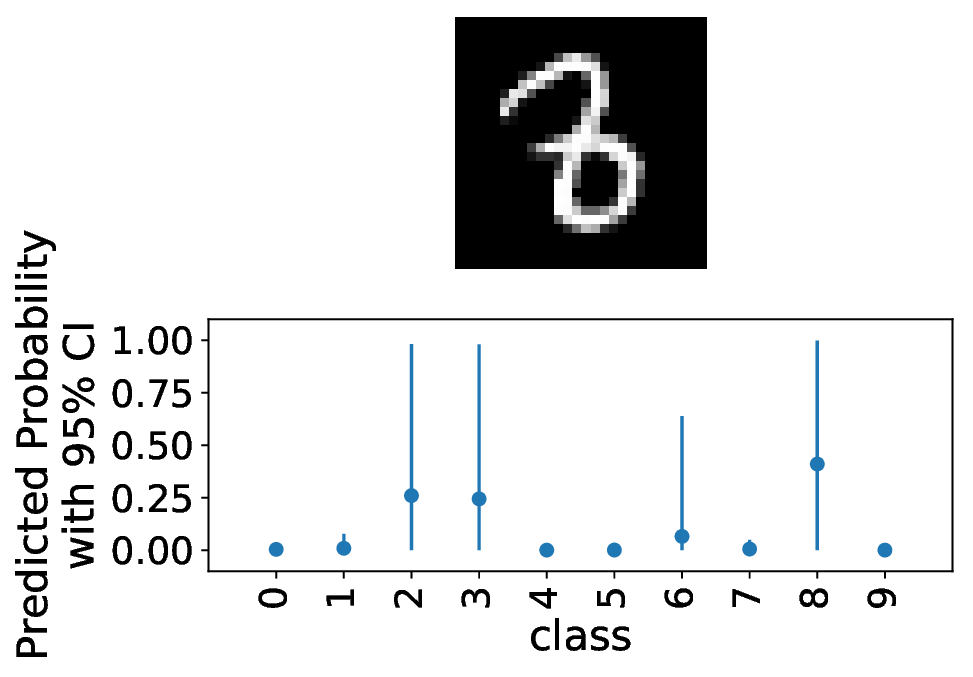}}
  \\
  \subfloat[][]{\includegraphics[width=0.33\linewidth]{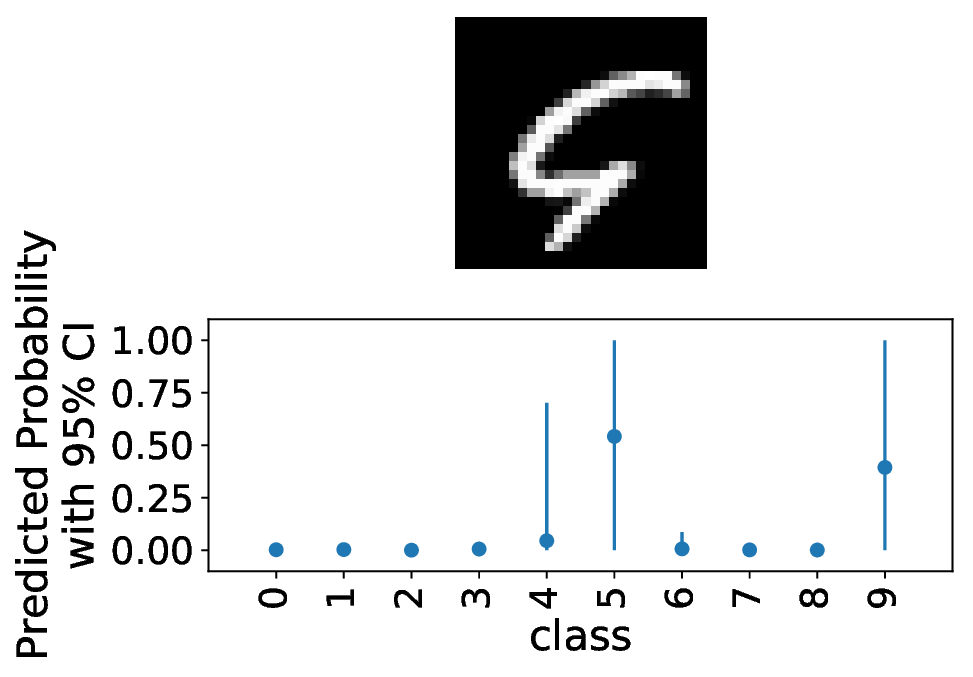}}
\caption{Examples of difficult to classify images from each class in MNIST. True class for each image is 0-9  arranged in alphabetical order. The bottom plot illustrates the 95\% credible intervals for these predictions. Best viewed on a computer screen.}
\end{figure*}

\end{document}